\documentclass{article}

\usepackage{spconf}
\usepackage{graphicx}
\usepackage{amsmath}
\usepackage{amssymb}
\usepackage{amsfonts}
\usepackage{subcaption}
\usepackage{color}

\usepackage[pagebackref=false,breaklinks=true,letterpaper=true,bookmarks=false]{hyperref}

\newlength{\subfigwidth}

\graphicspath{{figures/}}

\title{Learning-based Dequantization for Image Restoration Against
  Extremely Poor Illumination}

\name{Chang Liu$^{\dagger}$, \quad Xiaolin Wu$^{\dagger\ddagger}$,
  \quad Xiao Shu$^{\ddagger}$}
\address{$\dagger$ Shanghai Jiao Tong University\\
  $\ddagger$ McMaster University }

\begin{document}

\maketitle


\begin{abstract}
  All existing image enhancement methods, such as HDR tone mapping,
  cannot recover A/D quantization losses due to insufficient or
  excessive lighting, (underflow and overflow problems).  The loss of
  image details due to A/D quantization is complete and it cannot be
  recovered by traditional image processing methods, but the modern
  data-driven machine learning approach offers a much needed cure to
  the problem.  In this work we propose a novel approach to restore
  and enhance images acquired in low and uneven lighting.  First, the
  ill illumination is algorithmically compensated by emulating the
  effects of artificial supplementary lighting.  Then a DCNN trained
  using only synthetic data recovers the missing detail caused by
  quantization.
\end{abstract}

%
%
%
%
%
%

\section{Introduction}

Photographs taken in dark environments or in poor uneven illumination
conditions, such as in the night or backlighting, often become
illegible due to low intensity, much compressed dynamic range, low
contrast, and excessive noises.  Nowadays, even mass-marketed digital
cameras use high-resolution sensors and large capacity memory,
unsatisfactory spatial resolutions and compression artifacts are not
problems anymore.  But extremely poor lighting, which is beyond users'
control and defies autoexposure mechanism, remains a common,
uncorrected and yet understudied cause of image quality degradation.

The direct consequence of poor lighting is much compressed dynamic
range of the acquired image signal.  Existing image enhancement
methods can expand the dynamic range via tone mapping, but they are
inept to recover quality losses due to the A/D quantization of low
amplitude signals.  As a result, the tone-mapped images may appear
sufficiently bright with good contrast, but finer details are
completely erased. 

As the non-linear quantization operation is not invertible, the image
details erased by the A/D converter, when operating on weak and low
dynamic range image signals, cannot be recovered by traditional image
processing methods, such as high-pass filtering.  The technical
challenge in dequantizing images of compressed dynamic range is how to
estimate and compensate for the quantization distortions. Up to now
little has been done on the above missing data problem of A/D
dequantization, leaving consumers' long desire for low light cameras
unsatisfied.


In this work we propose a novel approach to restore and enhance images
acquired in low and uneven lighting.  First, the ill illumination is
algorithmically compensated by emulating the effects of artificial
supplementary lighting based on an image formation model.  The soft
light compensation is only an initial step to increase the overall
intensity and expand the dynamic range. It is not equivalent to
photographing using flash, for the quantization losses incurred in the
A/D conversion of low dynamic range images cannot be recovered in this
way.  Therefore, a subsequent step of the A/D dequantization is
required, and this task is particularly suited for the methodology of
deep learning, as will be demonstrated by this research.

Deep convolutional neural networks (DCNN) have been recently proven
highly successful in image restoration tasks, including
superresolution, denoising and inpainting.  But as the loss of
information due to quantization of low dynamic range images is not in
the spatial but pixel value domain, machine learning based A/D
dequantization appears to be more difficult than aforementioned other
problems of image restoration, and warrants some closer scrutiny.

As in all machine learning methods, the performance of the learnt
dequantization neural network primarily depends the quantity and
quality of the training data.  Using the same physical image formation
model for light compensation, we derive an algorithm to generate
training images of compressed dynamic range, by degrading
corresponding latent images of normal dynamic range (ground truth for
supervised learning).  The artifacts of the training images closely
mimic those caused by poor lighting conditions in real camera
shooting.  In addition to the good data quality, the generation
algorithm is designed in such a way that it can take ubiquitous,
widely available JPEG images as input, thus the machine learning for
the A/D dequantization task can benefit from practically unlimited
amount of training data.


%

\section{Related work}

As one of the fundamental problem in computer vision and image
processing, image enhancement has been widely used as a key step in
many applications such as image
classification~\cite{ciregan2012multi,yang2009linear}, image
recognition~\cite{he2016deep,simonyan2014very}, and object
tracking~\cite{yilmaz2006object}.  Many popular enhancement methods
are based on histogram
equalization~\cite{pisano1998contrast,abdullah2007dynamic}.  These
methods generally map the tone of the input image globally while
ignoring the relationship of pixels with their neighbors.  Variational
methods try to resolve this problem by imposing different
regularization terms based on different local features.  For instance,
contextual and variational contrast enhacement
(CVC)~\cite{celik2011contextual} finds the histogram mapping to get
large gray-level difference, while the method by Lee et
al.~\cite{lee2013contrast} enhances image by amplifying the gray-level
differences between adjacent pixels based on the layered difference
representation of 2D histogram.  Further more, optical tone mapping
(OCTM)~\cite{wu2011linear} was introduced for image enhancement via
optimal contrast-tone mapping.  Its variation~\cite{li2014contrast}
optimizes for maximal contrast gain while preserving the hue
component.

Another popular family of image enhancement algorithms is based on the
retinex theory, which explains the color and luminance perception
property of human vision system~\cite{land1977retinex}.  The most
important assumption of retinex theory is that an image can be
decomposed to illumination and reflectance.  Based on this idea,
single-scale retinex (SSR)~\cite{jobson1997properties} is designed to
estimate the reflectance and output it as an enhanced image.
Multi-scale retinex (MSR)~\cite{jobson1997multiscale} extends retinex
algorithm using multiple versions of the image on different scales.
Both SSR and MSR assume that the illumination image is spatially
smooth, which might not be true in real-world scenarios, as a result,
the output of these techniques often look unnatural in unevenly
illuminated regions.  LIME~\cite{guo2017lime} achieves good result by
imposing a structure prior on the illumination map.
SRIE~\cite{fu2016weighted} employs a weighted variational model to
estimate both the reflectance and the illumination, and apply this
model in manipulating the illumination map.  Based on the observation
that an inverted low-light image look like an image with haze, dehaze
techniques are also used for low-light image enhancement
~\cite{dong2011fast, li2015low}.  The work in~\cite{loza2013automatic}
is based on statistical modelling of wavelet coefficients of the
image.

Most existing low-light image enhancement techniques are model-based
rather than data-driven.  A recent neural network based attempt is
made to identify signal features from lowlight images and brighten
images adaptively without over-amplifying or saturating the lighter
parts in images with a high dynamic range~\cite{lore2017llnet}.
However, this technique only alleviates the uneven illumination,
leaving the problem of quantization unsolved.

\section{Preparation of Training Data}
\label{sec:data}

The proposed technique reduces the quantization artifacts of a
enhanced low-light image by using image details learned from other
natural images.  The effectiveness of our technique, or any machine
learning approaches, greatly relies on the availability of a
representative and sufficiently large set of training data.  In this
section, we discuss the methods for collecting and preparing the
training images for our technique.

To help the proposed technique identify the quantization artifacts
caused by poor illumination in real-world scenarios, ideally, the
training algorithm should only use real photographs as the training
data.  Obtaining a pair of low-light and normal images is easy; we can
take two consecutive shots of the same scene using different camera
exposure settings.  However, it is not easy to keep the pair of images
perfectly aligned, especially when the imaged subject, such as a
human, is in motion.  Another possible solution is to synthesize two
images of different brightness from a high dynamic range raw image by
simulating all the digital processes in a camera from demosaicing to
gamma correction to compression.  While image alignment is not a
concern using this method, collecting a large number of raw images of
various scenes is still a difficult and costly task.

In this research, we employ a simple data synthesis approach that
constructs realistic low-light images directly from normal low dynamic
range images.  As this approach does not require the original image to
be raw or high dynamic range, a huge number of images covering various
types of scenes are readily available online for training the purposed
technique.  To show how the data synthesis approach works, we first
assume that the formation of an image $I$ on camera can be modelled as
the piece-wise product of the illumination image $L$ and the
reflectance image $R$, as follows,
\begin{align}
  I(i) = L(i) R(i),
  \label{eq:I}
\end{align}
for a pixel at location $i$.

By the image formation model, if the illumination of the captured
scene decreases uniformly from $L$ to $a L$ by a factor of $a$ where
$a < 1$, then the captured image becomes attenuated to $a I$.
However, if an input image $J$ is in JPEG format, as do the majority
images available online, $J$ must have been gamma corrected and
quantized, i.e.,
\begin{align}
  J(i) = Q([I(i)]^{\gamma_1}),
  \label{eq:J}
\end{align}
where $\gamma_1$ is the gamma correction coefficient and
\begin{align}
  Q(x) = q \cdot \lfloor x / q + 0.5 \rfloor,
  \label{eq:Q}
\end{align}
is a quantization operator for some constant $q$.  As image $J$ is not
linear to the raw sensor reading $I$, simply multiplying $J$ by $a$
does not yield an accurate approximation of the corresponding
low-light image with light being dimmed by factor $a$.  Moreover, the
true sensor reading $I$ cannot be recovered by using inverse gamma
correction, as the gamma correction coefficient $\gamma_1$ is likely
unknown for image $J$ collected online.

Suppose $J_a$ is a JPEG image captured under dimmed light with factor
$a$, then by the definition of $I$ and $J$ in Eqs.~\eqref{eq:I} and
\eqref{eq:J},
\begin{align}
  J_a(i) &= Q([a L(i) R(i)]^{\gamma_2}), \nonumber
  \\
  &= Q(a^{\gamma_2} [I(i)]^{\gamma_2}),
  \label{eq:Ja}
\end{align}
where $\gamma_2$ is the gamma correction coefficient for the low-light
image, which is not necessarily the same as $\gamma_1$.  Since $J(i)$
is a quantized version of $[I(i)]^{\gamma_1}$ as defined in
Eq.~\eqref{eq:J},
\begin{align}
  J(i) = [I(i)]^{\gamma_1} + n_Q (i),
  \label{eq:Japprox}
\end{align}
where $n_Q(i)$ is the quantization noise.  Therefore, $J_a(i)$ can be
formulated as a function of $J(i)$, as follows,
\begin{align}
  J_a(i) & = Q(a^{\gamma_2} [J(i) - n_Q(i)]^{\gamma_2 / \gamma_1})
  \nonumber
  \\
  &= Q(a^{\gamma_2} [J(i)]^{\gamma_2 / \gamma_1}) + n(i),
  \label{eq:Ja2}
\end{align}
where term $n(i)$ accounts for the overall effects of $n_Q$ after
being gamma transformed and requantized.

Effectively, the low-light image $J_a$ is a gamma-corrected, dyanmic
range compressed and then quantized version of image $J$, as modelled
in Eq.~\eqref{eq:Ja2}.  By reverting the dynamic range compression and
gamma correction applied on $J_a$, we get a degraded image $\tilde{J}$
with correct exposure but tainted with realistic quantization
artifacts,
\begin{align}
  \tilde{J}(i) &= \left(\frac{J_a(i)}{a^{\gamma_2}}\right) ^ {\gamma_1 / \gamma_2}.
  \label{eq:Jt}
\end{align}
This is the way to generate a large, high-quality training data set
$\mathcal{T}$ to facilitate the deep learning method to be introduced
next.

\section{Quasi-$\ell_\infty$ Dequantization with Generative Adversary
  Neural Networks}

\subsection{Design Objective}

In order to solve the A/D dequantization problem, the standard method
of deep learning is to train a deep convolutional neural network $G$
that minimizes a loss function $L_G$, that is,
\begin{equation}
  G = \mathop{\arg\min}_{G} \sum_{(J, \tilde{J}) \in \mathcal{T}}
  L_G(J, G(\tilde{J})),
\end{equation}
where $(J, \tilde{J})$ is a sample pair drawn from $\mathcal{T}$ and
accordingly $\hat{J} = G(\tilde{J})$ is the input-output mapping of
network $G$.

But our problem has its unique characteristics, which need to be
reflected by the loss function $L_G$. First, the training image pairs
$(J, \tilde{J}) \in \mathcal{T}$ have a very high level of
variability, because we need to generate $\tilde{J} $
over a sufficiently large range of $a^{\gamma_2}$, $\gamma_2 /
\gamma_1$, and $n$.  This is necessary if the trained network $G$ is
to avoid the risk of data over fitting and perform robustly in all
poor lighting conditions and camera settings.  However, the spatial
structures of quantization residuals, which is the very information to
be recovered by deep learning, are largely independent of the lighting
level $a$ and parameters $\gamma_1$ and $\gamma_2$.  Therefore, we can
greatly reduce the variability of the outputs of $G$ and thus improve
the performance of the network, by changing the variables of the loss
function $L_G$:
\begin{equation}
  G = \mathop{\arg\min}_{G} \sum_{(J, \tilde{J}) \in \mathcal{T}} L_G(J-\tilde{J}, G(\tilde{J})).
  \label{eq:G}
\end{equation}
In other words, network $G$ learns to predict, from $\tilde{J}$, the
quantization residual,
\begin{equation}
  E(\tilde{J}) = J-\tilde{J},
\end{equation}
rather than the latent image $J$ directly.

\begin{figure*}[t]
  \centering
  \includegraphics[width=0.8\linewidth]{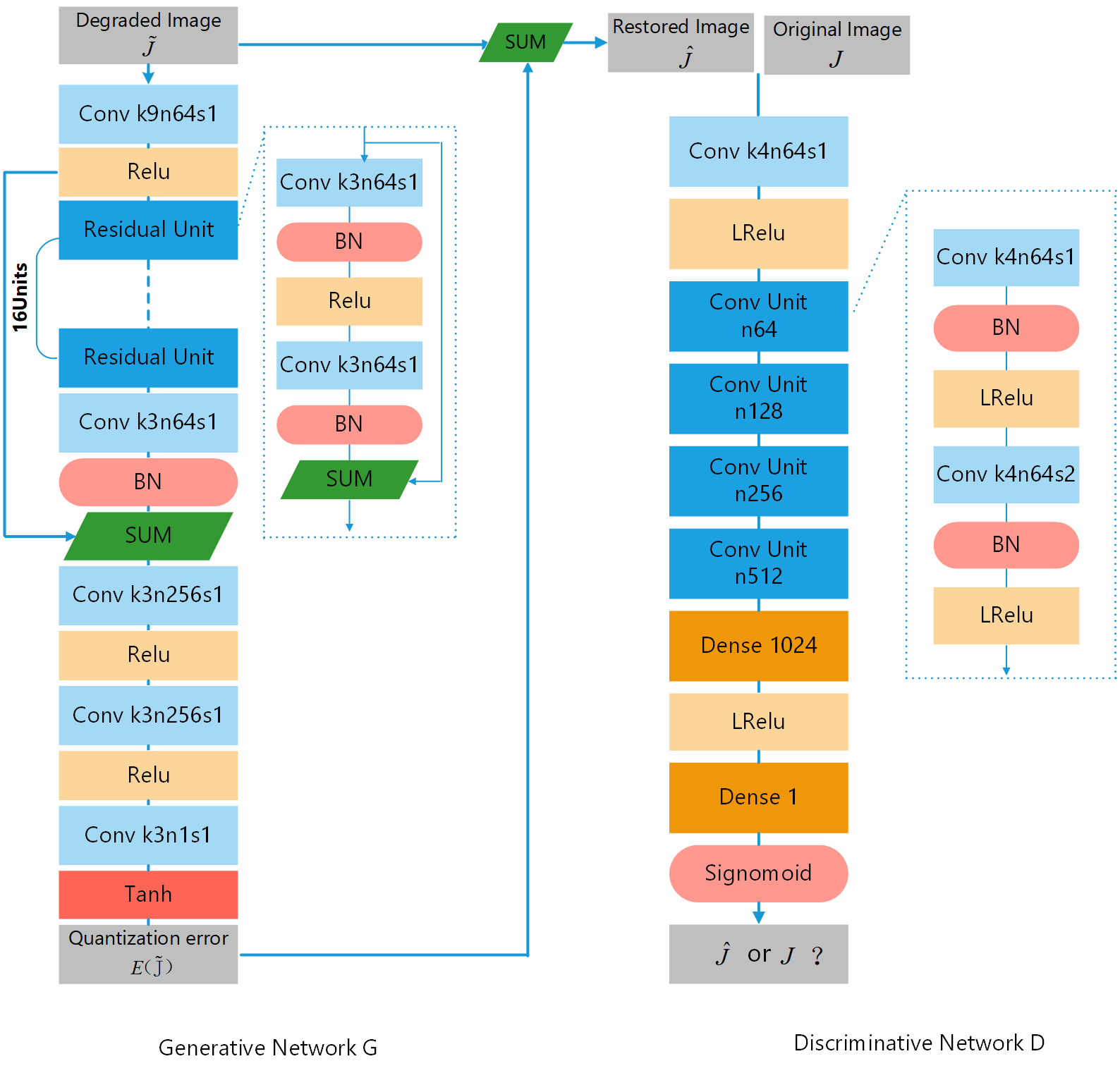}
  \caption{The architecture of the proposed generative and adversarial
    networks.}
  \label{fig:netG}
\end{figure*}

\subsection{GAN Construction}

The next critical design decision is what are suitable quality
criteria for the reconstructed image $\hat{J}$, which is to guide the
construction of network $G$.  For most users, the goals of enhancing
poorly exposed images are overall legibility and aesthetics; signal
level precision is secondary.  This image quality preference is
exactly the strength of the adversarial neural networks (GAN).  In
GAN, two neural networks, one called the generative network $G$ and
the other the discriminative network $D$, contest against each other.
In our case, network $G$ is the one stated in Eq.~\eqref{eq:G}.  It
strives to generate an image $G(\tilde{J})$ to past the test of being
a properly exposed image $J$ that is conducted by network $D$.  On the
other hand, network $D$ is trained to discriminate and reject the
output images of network $G$.

The two competing networks $G$ and $D$ are constructed as shown in
Fig.~\ref{fig:netG}.  The image $\tilde{J}$ of poor exposure and
compressed dynamic range is fed into the generative network $G$ to be
repaired.  Network $G$ is trained to predict the quantization
residual.  Adding this residual to input image $\tilde{J}$ yields a
restored image $\hat{J}$.  Then, the restored image $\hat{J}$ and the
latent image $J$ are used to train the discriminative network $D$.
Reciprocally, network $D$ outputs its discriminator result of
$\tilde{J}$ to help train network $G$.

As illustrated in Fig.~\ref{fig:netG}, network $G$ is constructed as a
deep convolution neural network, to exploit its ability to learn a
mapping of high complexity.
Our network $G$ contains 16 residual units~\cite{he2016deep}, each
consisting of two convolutional layers, two batch normalization
layers~\cite{ioffe2015batch} and one ReLU activation layers. For the
architecture of network $D$, we borrowed the design of
DCGAN~\cite{radford2015unsupervised}. It has four convolution units,
each of which has two convolution layers, one with stride 1 and the
other with stride 2.  They are respectively followed by one batch
normalization layer and one leakyRelu activation layer ($\alpha =
0.2$)~\cite{nair2010rectified}.




The output of the discriminative network, $D(\hat{J})$ or $D(J)$, can
be interpreted as the probability that the underlying image is
acquired in proper illumination conditions.

Following the idea of Goodfellow {\it et al.}, we set a discriminative
network $D$, which is optimized in conjunction with $G$, to solve the
following min-max problem:
\begin{align}
  \min_G \max_D \left\{ \mathbb{E}_{J} \big[ \log D(J) \big] +
  \mathbb{E}_{\hat{J}} \big[ \log(1-D(G(\hat{J}))) \big] \right\}
\end{align}

In practice, for better gradient behavior, we minimize $-\log
D(G(\hat{J}))$ instead of $\log(1-D(G(\hat{J})))$, as proposed in
\cite{goodfellow2014generative}.  This introduces an adversarial term
in the loss function of the generator CNN $G$:
\begin{align}
  L_\oslash = -\log D(G(\hat{J})) .
\end{align}
In competition against generator network $G$, the loss function for
training discriminative network $D$ is the binary cross entropy:
\begin{align}
L_D = - \big[ \log(D(J))+\log(1-D(G(\hat{J}))) \big]
\label{loss_D}
\end{align}
Minimizing $L_\oslash$ drives network $G$ to produce restored images
that network $D$ cannot distinguish from properly exposed images.
Accompanying the evolution of $G$, minimizing $L_D$ increases the
discrimination power of network $D$.


%
%

%
%

\subsection{Quasi-$\ell_\infty$ Loss}

As observed in ~\cite{ledig2016photo}~\cite{guo2016one}, adversarial
training will drive the reconstruction towards the natural image
manifold, producing perceptually agreeable results. However, the
generated images are prone to fabricated structures that deviate too
much from the ground truth.  Particularly in our case, the learnt
quantization residuals are to be added onto a base layer image. If
these added structures are unrestricted at all, they could cause
undesired artifacts, such as halos.
To overcome these weaknesses of adversarial training, we introduce a
structure-preserving quasi-$\ell_\infty$ loss term $L_\infty$ to
tighten up the signal-level slack of probability divergence loss terms
$L_D$ and $L_\oslash$.


To construct the loss term $L_\infty$, we first show that the restored
image should be bounded by the degraded image $\tilde{J}$.  By the
definitions of $\tilde{J}$ in Eq.~\eqref{eq:Jt} and the low-light
image $J_a$  in Eq.~\eqref{eq:Ja2}, we have,
\begin{align}
  Q(a^{\gamma_2} [J(i)]^{\gamma_2 / \gamma_1}) + n(i) &= a ^ {\gamma_2} [\tilde{J}(i)]^{\gamma_2 / \gamma_1}.
\end{align}
Then, by the definition of the quantization operator $Q(x)$ in
Eq.~\eqref{eq:Q},
\begin{align}
  \left| a^{\gamma_2} [\tilde{J}(i)]^{\gamma_2/\gamma_1}-n(i) - a^{\gamma_2} [J(i)]^{\gamma_2 / \gamma_1} \right|
  & \le \frac{q}{2}.
\end{align}



\begin{figure}[t]
  \centering
  \includegraphics[width=0.9\linewidth]{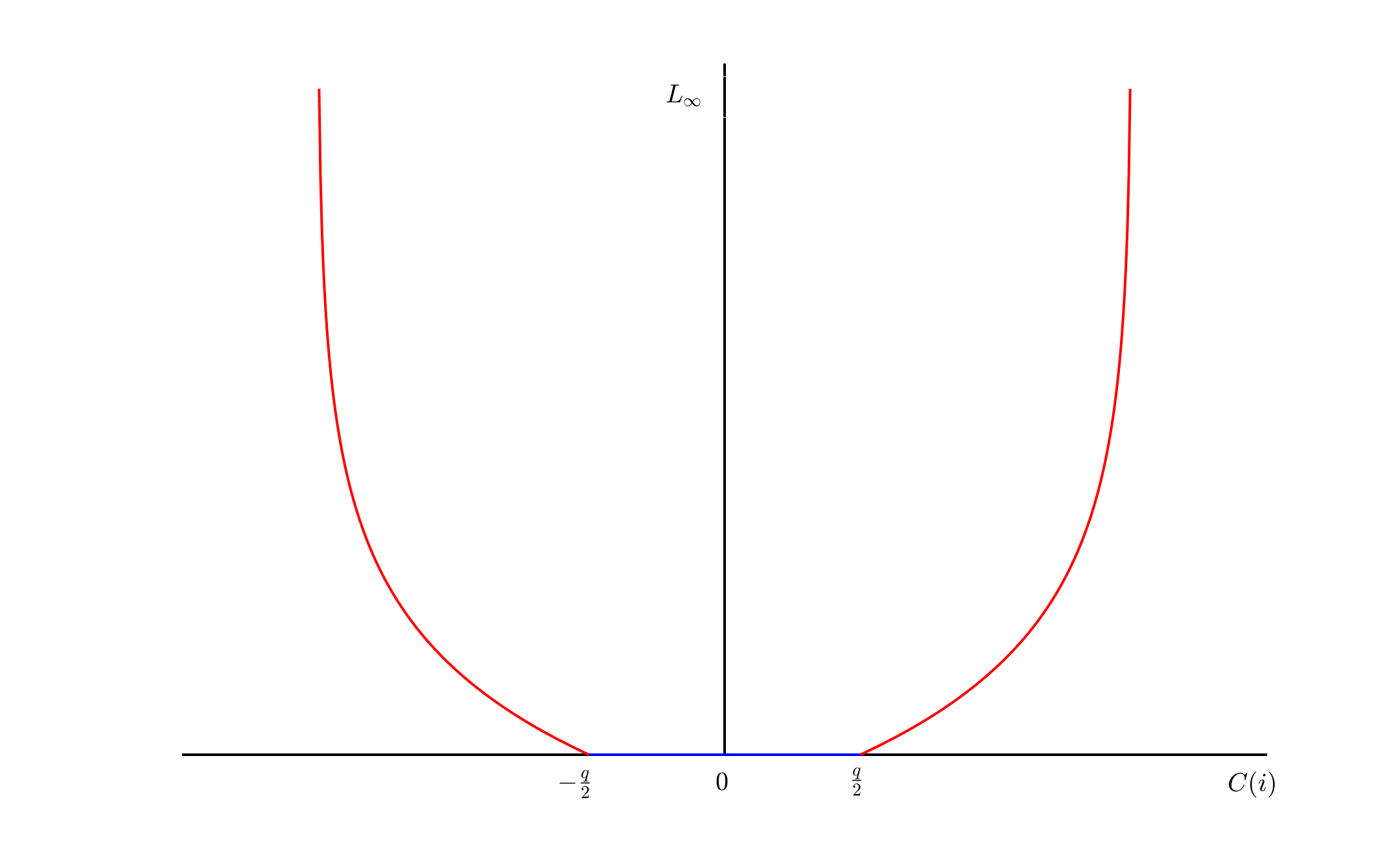}
  \caption{The illustration of $L_{\infty}$ loss for one pixel}
  \label{fig:infty}
\end{figure}

To enforce these inequalities in the proposed neural network, we
employ a barrier function in the quasi-$l^\infty$ loss function
$L_{\infty}$ as follows,
\begin{equation}
  L_{\infty} = \sum_{i \in B} -\log \left( 1 - \max\left\{|C(i)|-\frac{q}{2},0\right\}\right),
\end{equation}
where $B$ is the pixel patch used in the training and,
\begin{align}
  C(i) = a^{\gamma_2} [\tilde{J}(i)]^{\gamma_2/\gamma_1}- a^{\gamma_2} [\hat{J}(i)]^{\gamma_2 / \gamma_1} - n(i).
\end{align}
Plotted in Fig.~\ref{fig:infty} is the quasi-$l^\infty$ loss function.
As shown in the figure, the quasi-$\ell_\infty$ loss is 0 within the
quantization interval, and it increases rapidly once the pixel value
falls outside of the interval.



Finally, we combine the adversarial loss and quasi-$\ell_\infty$ loss
when optimizing the generative network $G$, namely,
\begin{equation}
  L_G = L_{\infty} + \lambda L_\oslash.
  \label{eq:joint training}
\end{equation}

\section{Locally Adaptive Illumination Compensation}
\label{sec:tone}

As discussed in Section~\ref{sec:data}, the proposed DCNN technique
learn the pattern of quantization artifacts from dynamic range
stretched image patches generated using Eq.~\eqref{eq:Jt}.  Ideally,
the proposed technique should work the best if the input image is also
stretched by such a simple linear tone mapping.  However, linear tone
mapping, which adjusts the illumination of an image uniformly, is too
restrictive in practice.  For any image with a wide dynamic range,
such as a photo containing both underexposed and normally exposed
regions, linear tone mapping cannot enhance the dark regions
sufficiently without saturating the details in the bright regions.
Thus, it is necessary to adopt a locally adaptive approach for
compensating the illumination of the input image.

There are plenty of tone mapping operators that can adjust image
brightness locally~\cite{zfarbman2008, rfattal2002, sparis2011,
  zrahman2004, ereinhard2002}, but none of these existing techniques
fit all the requirements of the proposed approach.  For an image with
severely underexposed regions, the proposed dequantization neural
network needs each of these regions to be stretched as uniformly as
possible, just like a uniform increase of illumination as modelled in
Eq.~\eqref{eq:I}.  Additionally, the tone mapped image should also
exhibit good contrast, making local detail more visible to human
viewers.  Combining these two requirements together results a new
formulation for low-light or uneven-light image tone mapping, namely
locally adaptive illumination compensation (LAIC) as follows,
\begin{equation}
  \begin{array}{rl}
    \mbox{minimize} & \displaystyle\sum_{i=1}
    \left[ D \left(
        \tilde{J}(i) / J_a(i)
      \right)
      - \lambda_2 \left| \tilde{J}(i) - \bar{J}(i) \right| \right] \\
    \mbox{subject to}
    & 0 \le \tilde{J}(i) \le 1, \\
    & \operatorname{sgn}\left(\tilde{J}(i) - \bar{J}(i)\right)
    = \operatorname{sgn}\left(J_a(i) - \bar{J}_a(i)\right),
  \end{array}
  \label{eq:tone}
\end{equation}
where the enhanced image $\tilde{J}$ is the variable, and the original
low-light image $J_a$ is constant to the problem.  Operator
$\operatorname{sgn}(\cdot)$ is the sign function.  Variable
$\bar{J}(i)$ represents the average pixel intensity of image
$\tilde{J}$ in the neighbourhood $\mathcal{D}_i$ of pixel $i$, i.e.,
\begin{equation}
  \bar{J}(i) = \frac{1}{|\mathcal{D}_i|} \sum_{k \in \mathcal{D}_i} \tilde{J}(k).
\end{equation}
Similarly, $\bar{J}_a(i)$ is the local average of $J_a(i)$.  For
a pixel of image $I$ with coordinate $(u,v)$, derivative operator
$D\left(I(u,v)\right)$ is defined as follows,
\begin{align}
  D\left(I(u,v)\right) &= |I(u + 1,v)-I(u,v)|
  \nonumber
  \\
  &\quad + |I(u, v+1)-I(u,v)|.
\end{align}

The problem in Eq.~\eqref{eq:tone} optimizes two objectives: the total
variation of illumination gain and the local contrast of the enhanced
image.  Minimizing the total variation of illumination gain
$\tilde{J}(i) / J_a(i)$ is to find a solution with piece-wise constant
illumination gain.  Maximizing the local contrast is to boost the
detail of the output image.  The importance of the two often
conflicting objectives are balanced with a user given Lagrange
multiplier $\lambda_2$.

The first constraint of the optimization problem in
Eq.~\eqref{eq:tone} is to bound the dynamic range of the enhanced
image $\tilde{J}$ to $[0,1]$.  The second constraint is to preserve
the rank of each pixel to its local average.  For instance, if a pixel
of the input image is brighter than the average pixel value in the
neighbourhood of the pixel, then the same must also be true in the
enhanced output image by this constraint.  By preserving the rank in
local regions, a method can be perceptually free of many tone mapping
artifacts such as Halo and double edge.

This optimization problem is non-convex and difficult to solve
directly, however, since the enhanced image $\tilde{J}$ preserves the
rank between each pixel and its local average, $\tilde{J}(i) -
\bar{J}(i)$ always shares the same sign with $J_a(i) -
\bar{J}_a(i)$.  Thus,
\begin{align}
  \left| \tilde{J}(i) \! - \! \bar{J}(i) \right|
  &= \left( \tilde{J}(i) \! - \! \bar{J}(i) \right) \cdot
  \operatorname{sgn}\left(\tilde{J}(i) \! - \! \bar{J}(i) \right)
  \nonumber
  \\
  &= \left( \tilde{J}(i) \! - \! \bar{J}(i) \right) \cdot
  \operatorname{sgn}\left(J_a(i) \! - \! \bar{J}_a) \right)
  \label{eq:objective}
\end{align}
Since $J_a(i)$ and $\bar{J}_a(i)$ are constant to the
optimization problem in Eq.~\eqref{eq:tone}, the local contrast term
of the objective function is linear.  On the other hand, the total
variation of illumination gain term can also be reformulate as a
linear function.  Thus, the objective function of the problem is
linear.  The local rank preserving constraints in Eq.~\ref{eq:tone}
can also be written as equivalent linear inequalities as follows,
\begin{equation}
  \left\{
    \begin{array}{rr}
      \tilde{J}(i) - \bar{J}(i) \le 0 & \text{if $J_a(i) \le \bar{J}_a(i)$}, \\
      \tilde{J}(i) - \bar{J}(i) \ge 0 & \text{if $J_a(i) \ge \bar{J}_a(i)$}. \\
    \end{array}
  \right.
  \label{eq:preserve2}
\end{equation}
Therefore, this reformulated LAIC problem is a tractable linear
program.

\section{Experimental Results}

In this section, We compare our method with four of the
state-of-the-art low-light image enhancement methods: CLAHE
\cite{pisano1998contrast}, OCTM \cite{wu2011linear}, LIME
\cite{guo2017lime} and SRIE \cite{fu2016weighted}.  The operation of
CLAHE is executed on the V channel of images by first converting it
from RGB colorspace to the HSV colorspace and then converting the
proposed HSV back to the RGB colorspace.  LLNeT \cite{lore2017llnet}
is not examined here, as the authors did not make the implementation
available for evaluation.  We evaluate the tested methods on variety
of data, including synthetic images, high dynamic range images and
real photographs.  Except for the first experiment on the synthetic
global low-light images, we firstly enhance the images by LAIC, then
restore the quantization residual using our trained network.

\subsection{Experiments on Synthetic Images}

\begin{figure*}[t]
  \centering

  \setlength{\subfigwidth}{0.16\linewidth}

  \begin{subfigure}[b]{\subfigwidth}
    \includegraphics[width=\textwidth, trim=0 50 0 50, clip]{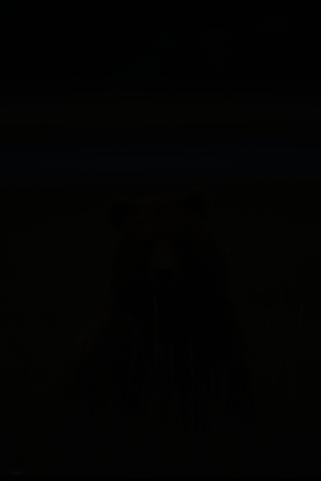}
    \\[2pt]
    \includegraphics[width=\textwidth, trim=50 0 50 0, clip]{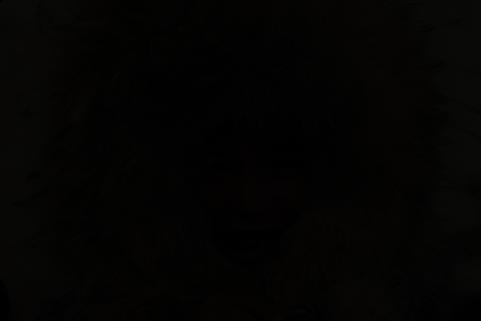}
    \caption{Input}
  \end{subfigure}
  \begin{subfigure}[b]{\subfigwidth}
    \includegraphics[width=\textwidth, trim=0 50 0 50, clip]{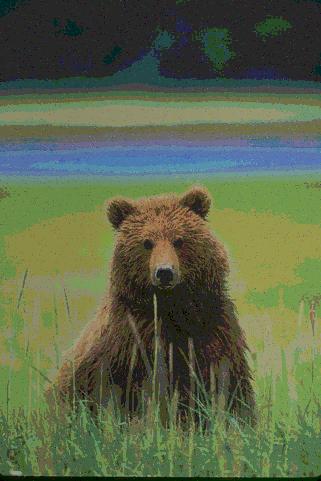}
    \\[2pt]
    \includegraphics[width=\textwidth, trim=50 0 50 0, clip]{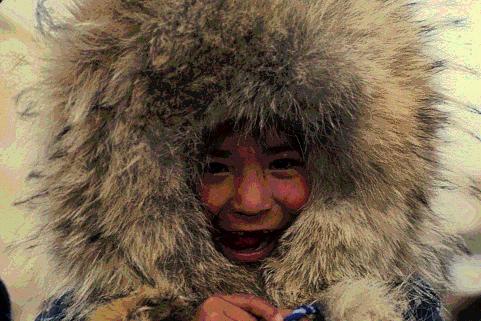}
    \caption{CLAHE}
  \end{subfigure}
  \begin{subfigure}[b]{\subfigwidth}
    \includegraphics[width=\textwidth, trim=0 50 0 50, clip]{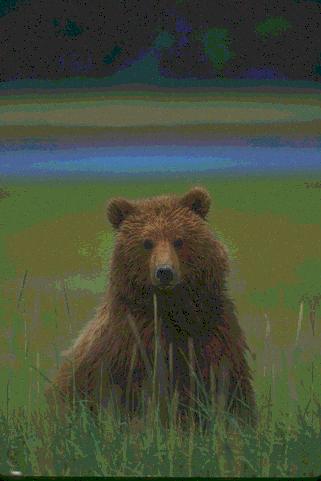}
    \\[2pt]
    \includegraphics[width=\textwidth, trim=50 0 50 0, clip]{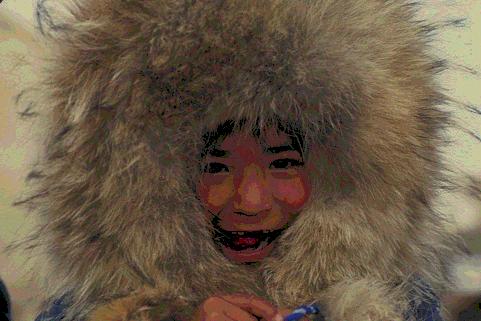}
    \caption{OCTM}
  \end{subfigure}
  \begin{subfigure}[b]{\subfigwidth}
    \includegraphics[width=\textwidth, trim=0 50 0 50, clip]{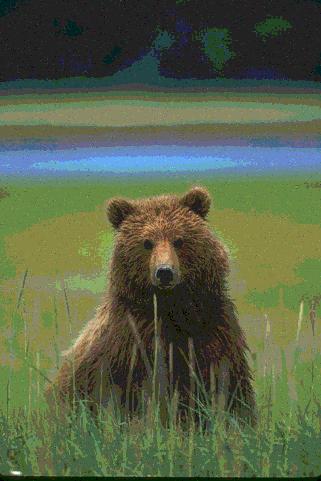}
    \\[2pt]
    \includegraphics[width=\textwidth, trim=50 0 50 0, clip]{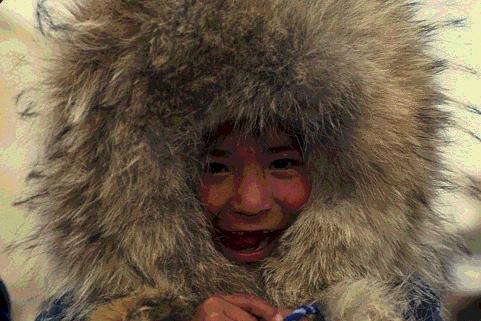}
    \caption{LIME}
  \end{subfigure}
  \begin{subfigure}[b]{\subfigwidth}
    \includegraphics[width=\textwidth, trim=0 50 0 50, clip]{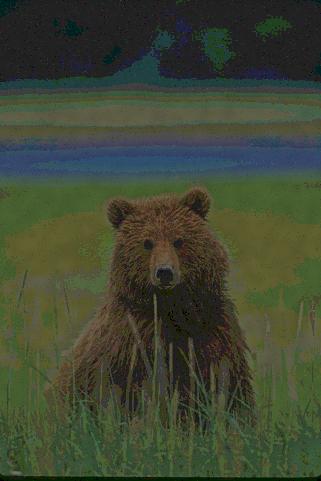}
    \\[2pt]
    \includegraphics[width=\textwidth, trim=50 0 50 0, clip]{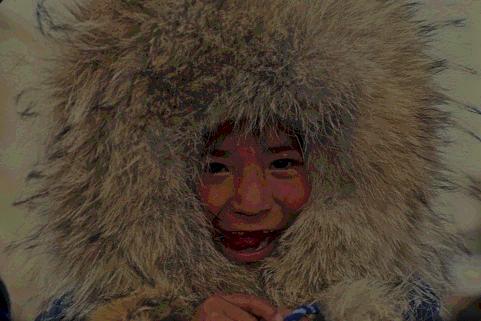}
    \caption{SRIE}
  \end{subfigure}
  \begin{subfigure}[b]{\subfigwidth}
    \includegraphics[width=\textwidth, trim=0 50 0 50, clip]{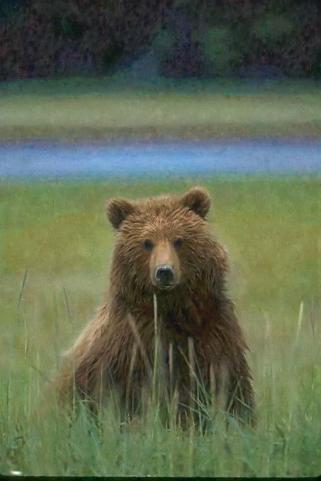}
    \\[2pt]
    \includegraphics[width=\textwidth, trim=50 0 50 0, clip]{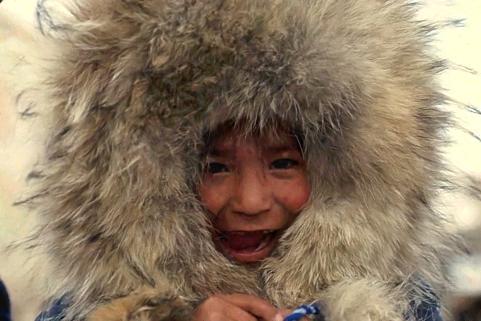}
    \caption{Proposed}
  \end{subfigure}
  \caption{Results of tested techniques for synthetic low-light
    images.  In this experiment, the low-light images are linearly
    tone mapped before applying the proposed dequantization
    technique.}
  \label{fig:syn-global}
\end{figure*}

\begin{figure*}[t]
  \centering

  \setlength{\subfigwidth}{0.16\linewidth}

  \begin{subfigure}[b]{\subfigwidth}
    \includegraphics[width=\textwidth, trim=50 0 50 0, clip]{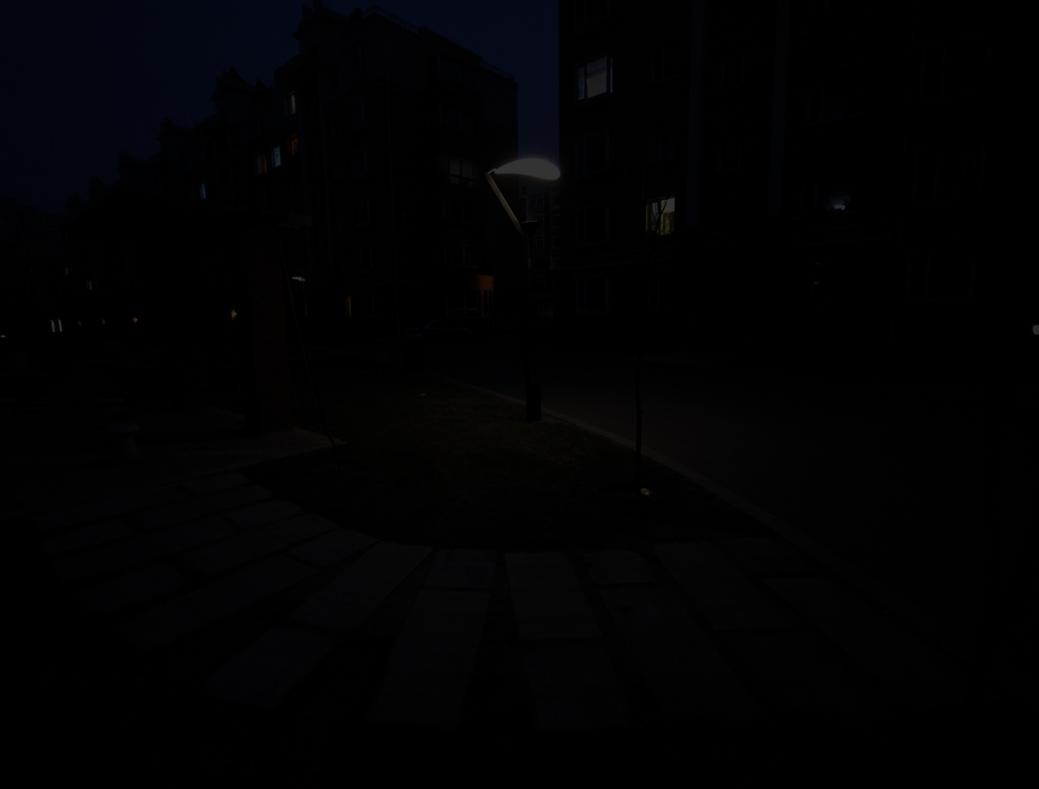}
    \\[2pt]
    \includegraphics[width=\textwidth, trim=0 0 0 0, clip]{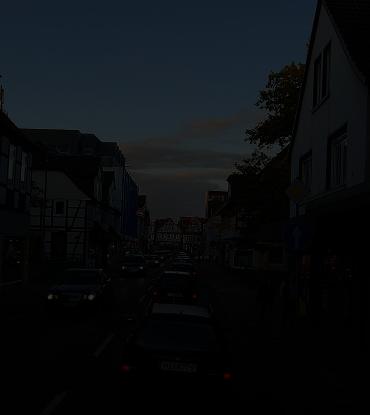}
    \caption{Input}
  \end{subfigure}
  \begin{subfigure}[b]{\subfigwidth}
    \includegraphics[width=\textwidth, trim=50 0 50 0, clip]{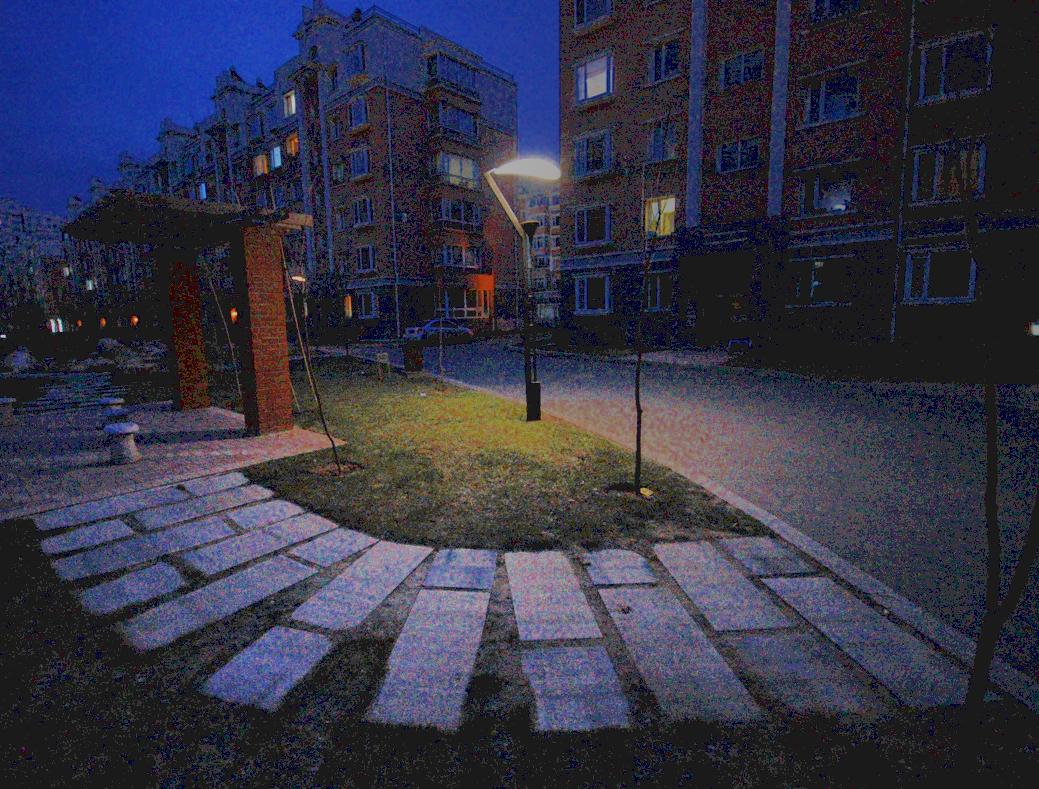}
    \\[2pt]
    \includegraphics[width=\textwidth, trim=0 0 0 0, clip]{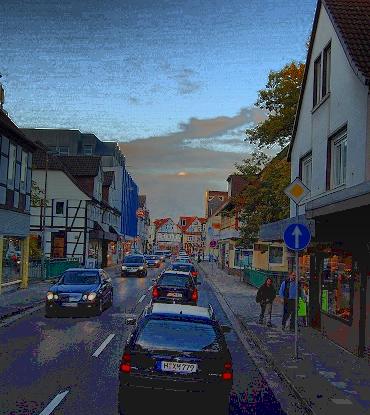}
    \caption{CLAHE}
  \end{subfigure}
  \begin{subfigure}[b]{\subfigwidth}
    \includegraphics[width=\textwidth, trim=50 0 50 0, clip]{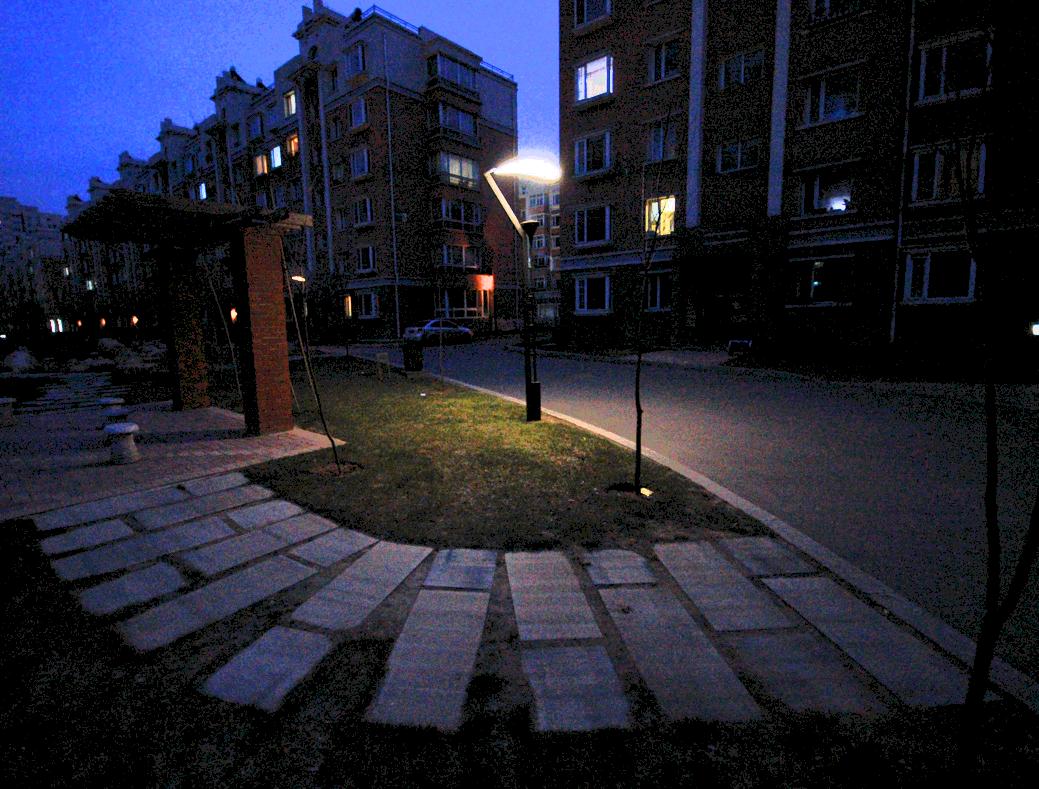}
    \\[2pt]
    \includegraphics[width=\textwidth, trim=0 0 0 0, clip]{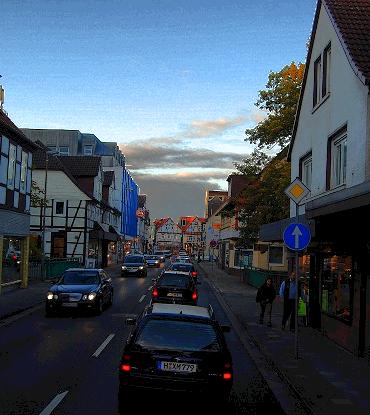}
    \caption{OCTM}
  \end{subfigure}
  \begin{subfigure}[b]{\subfigwidth}
    \includegraphics[width=\textwidth, trim=50 0 50 0, clip]{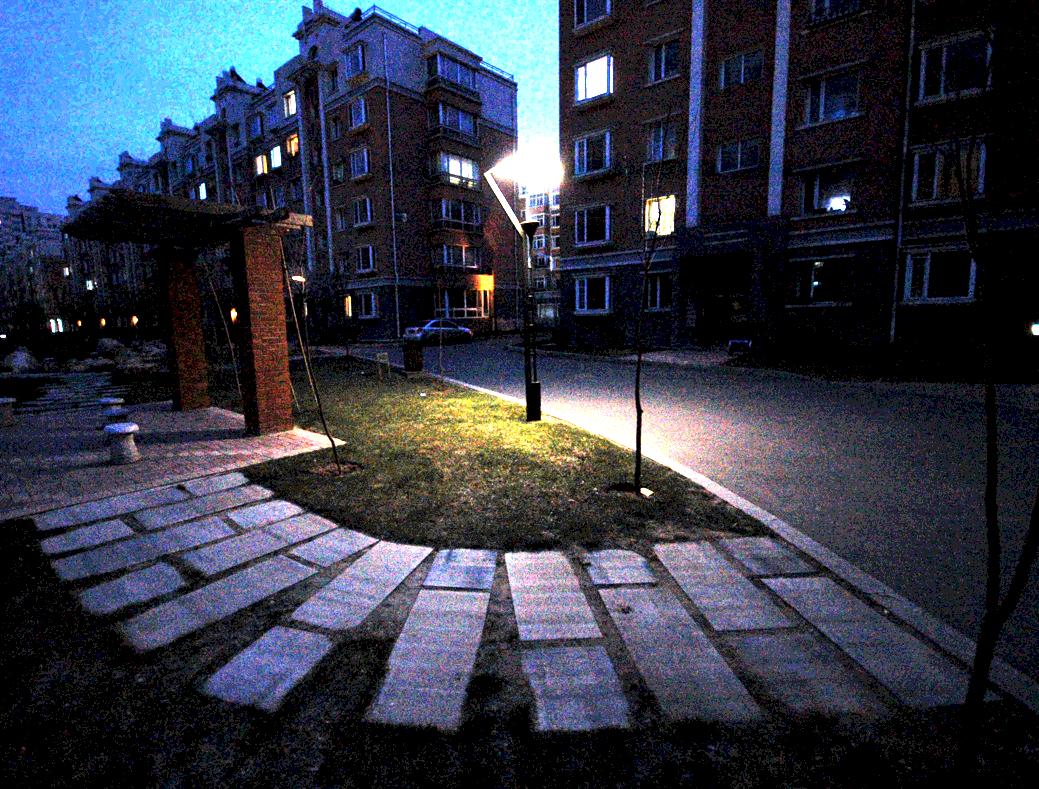}
    \\[2pt]
    \includegraphics[width=\textwidth, trim=0 0 0 0, clip]{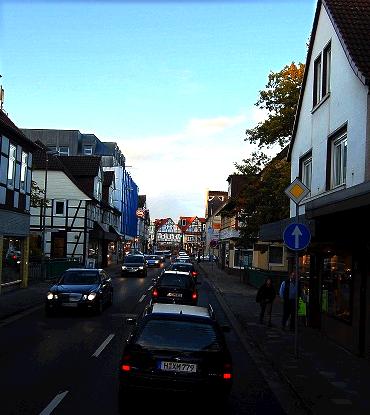}
    \caption{LIME}
  \end{subfigure}
  \begin{subfigure}[b]{\subfigwidth}
    \includegraphics[width=\textwidth, trim=50 0 50 0, clip]{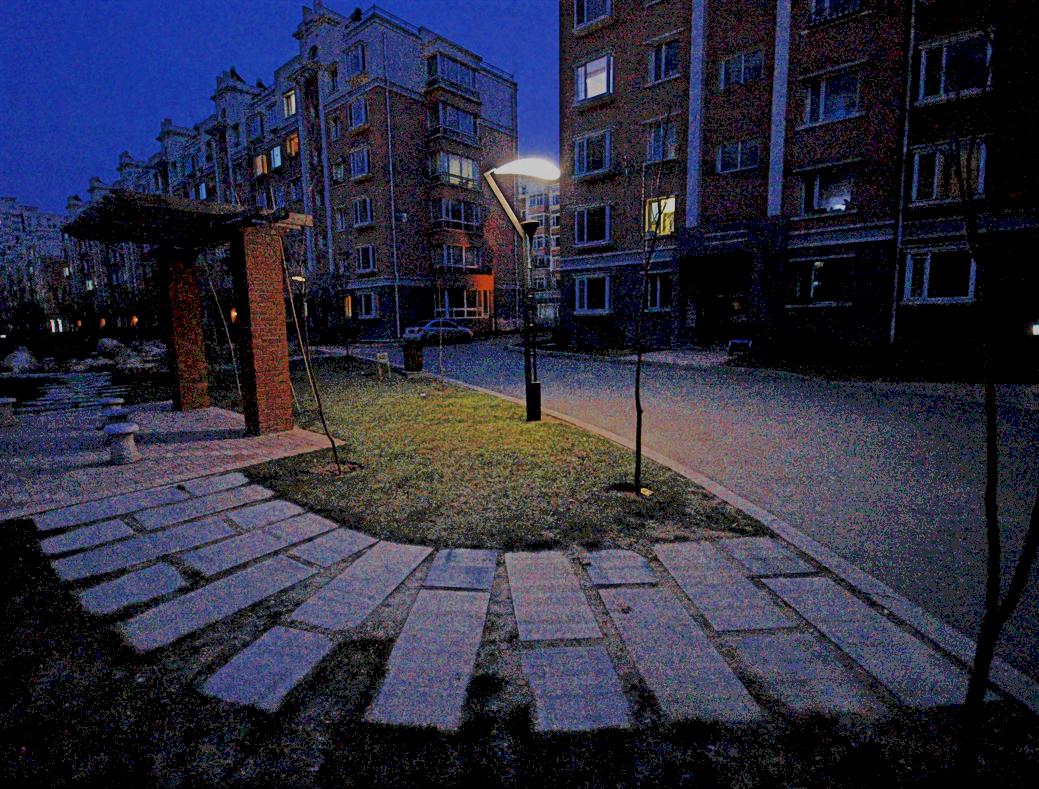}
    \\[2pt]
    \includegraphics[width=\textwidth, trim=0 0 0 0, clip]{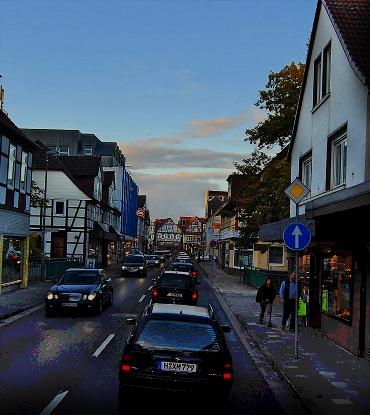}
    \caption{SRIE}
  \end{subfigure}
  \begin{subfigure}[b]{\subfigwidth}
    \includegraphics[width=\textwidth, trim=50 0 50 0, clip]{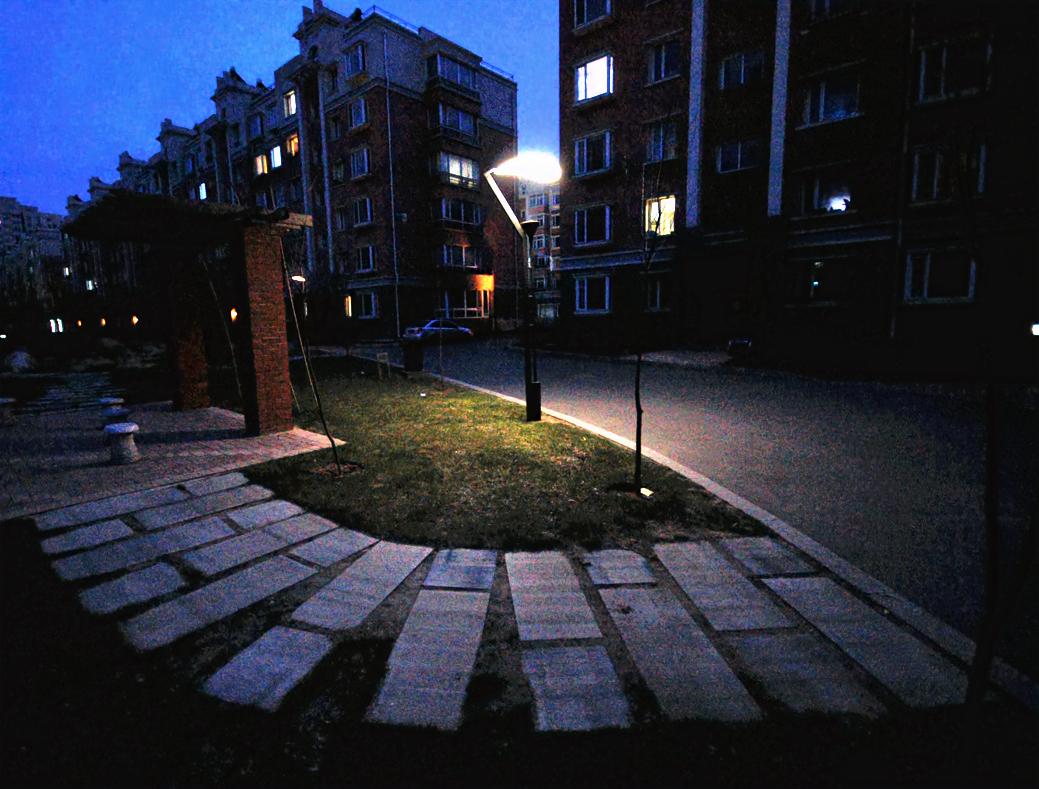}
    \\[2pt]
    \includegraphics[width=\textwidth, trim=0 0 0 0, clip]{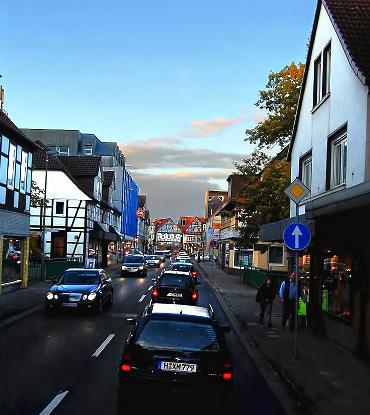}
    \caption{Proposed}
  \end{subfigure}
  \caption{Results of the tested techniques for synthetic low-light
    images.  In this experiment, the low-light images are tone mapped
    by LAIC before applying the proposed dequantization technique.}
  \label{fig:syn-local}
\end{figure*}

\begin{figure*}[t]
  \centering

  \setlength{\subfigwidth}{0.16\linewidth}

  \begin{subfigure}[b]{\subfigwidth}
    \includegraphics[width=\textwidth, trim=0 0 0 0, clip]{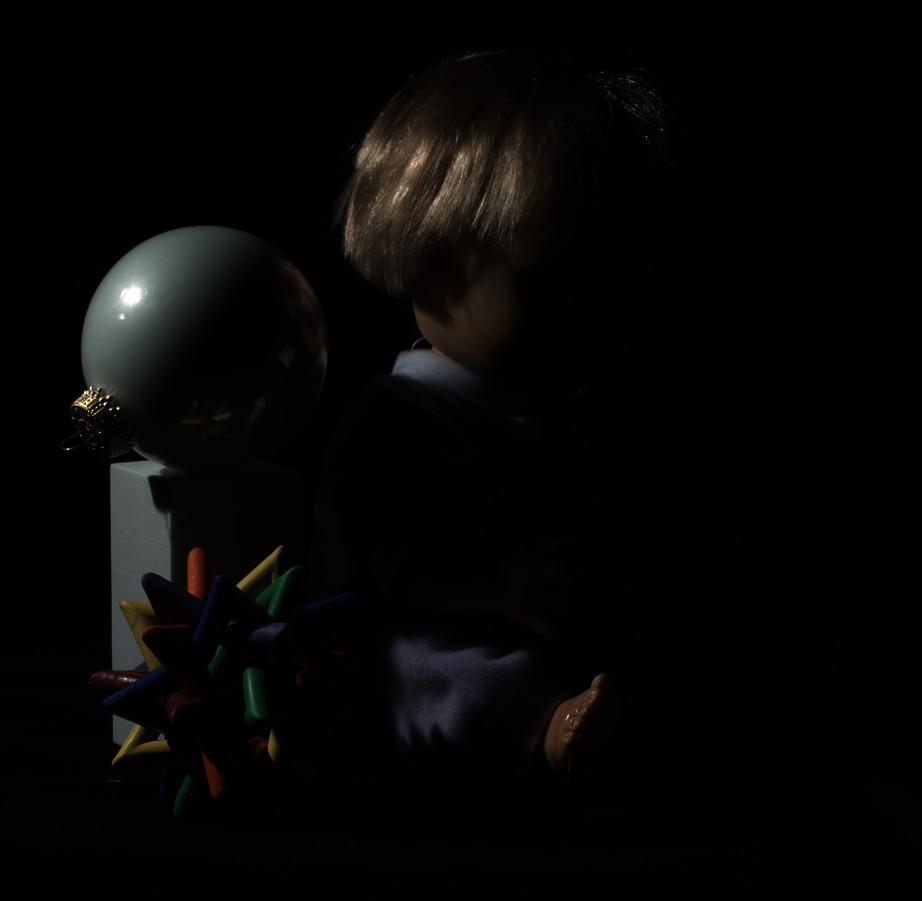}
    \\[2pt]
    \includegraphics[width=\textwidth, trim=50 0 150 0, clip]{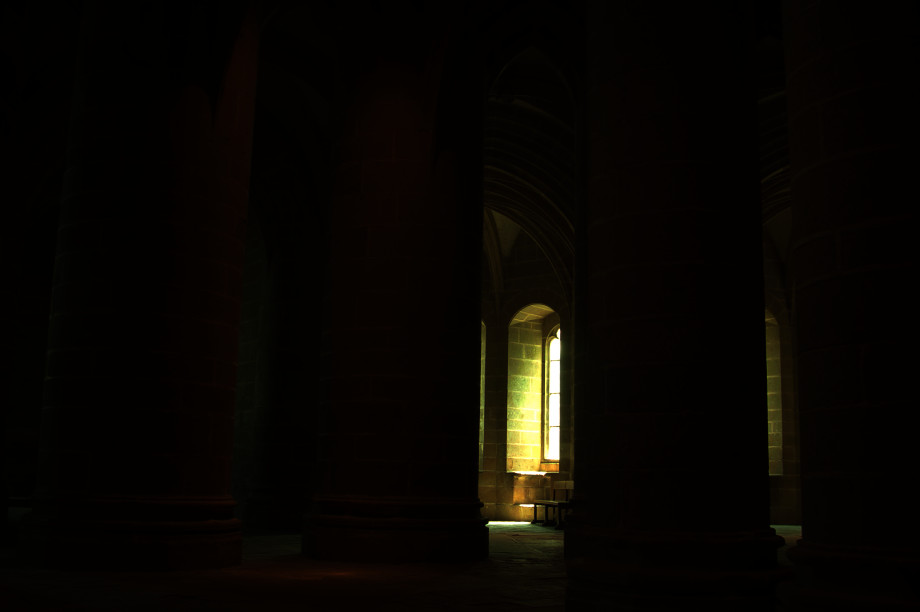}
    \\[2pt]
    \includegraphics[width=\textwidth, trim=0 150 0 0, clip]{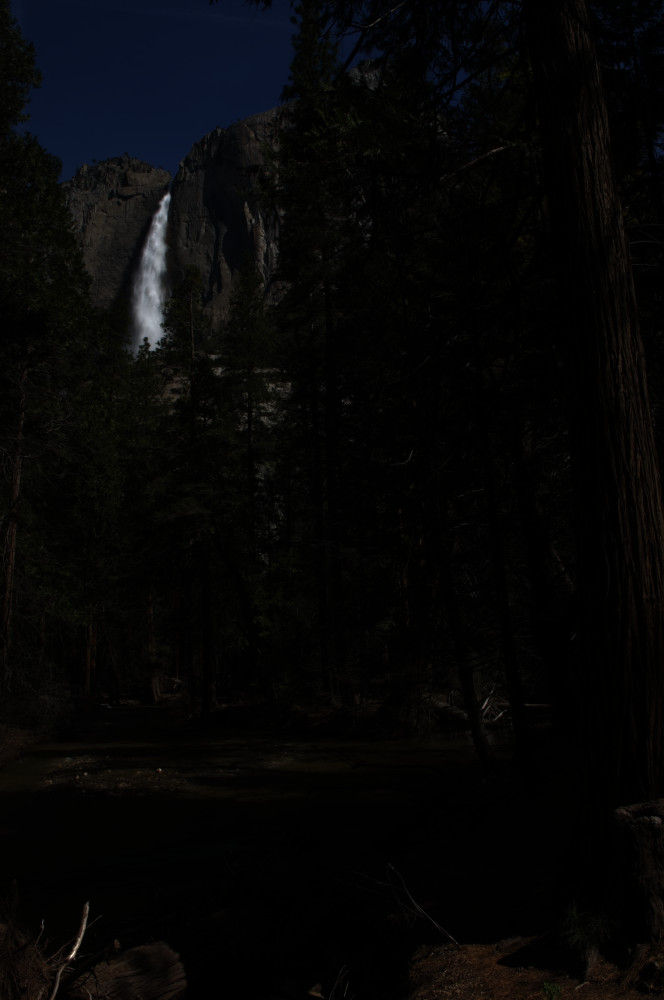}
    \caption{Input}
  \end{subfigure}
  \begin{subfigure}[b]{\subfigwidth}
    \includegraphics[width=\textwidth, trim=0 0 0 0, clip]{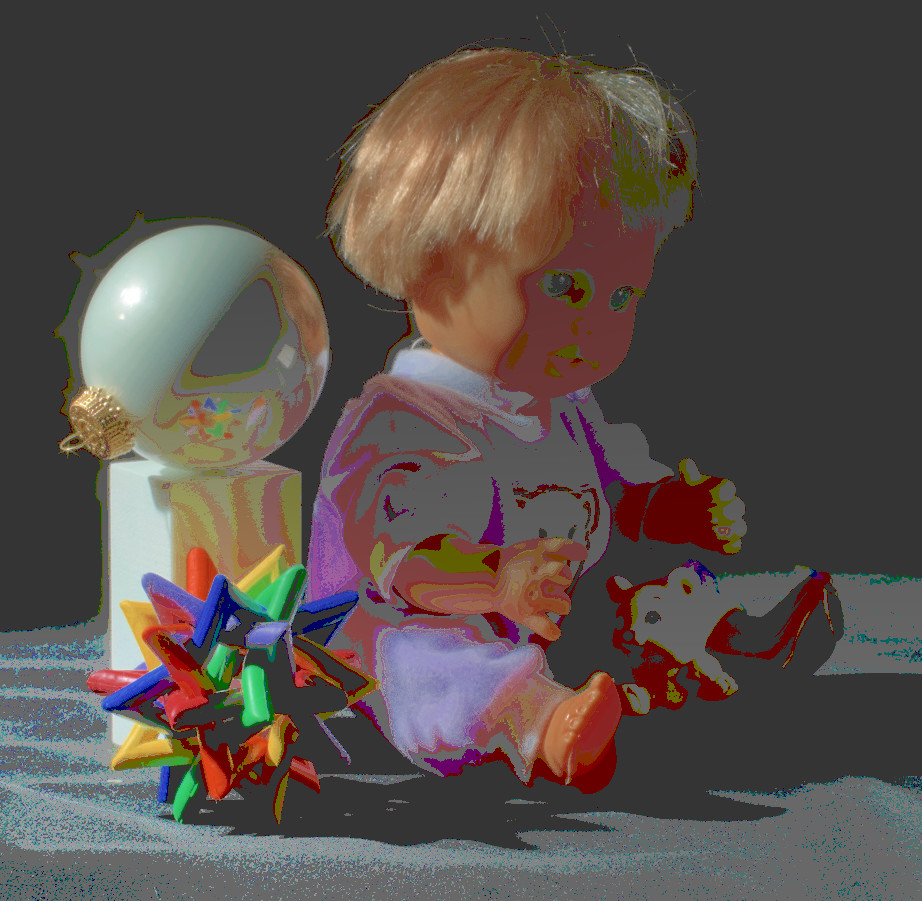}
    \\[2pt]
    \includegraphics[width=\textwidth, trim=50 0 150 0, clip]{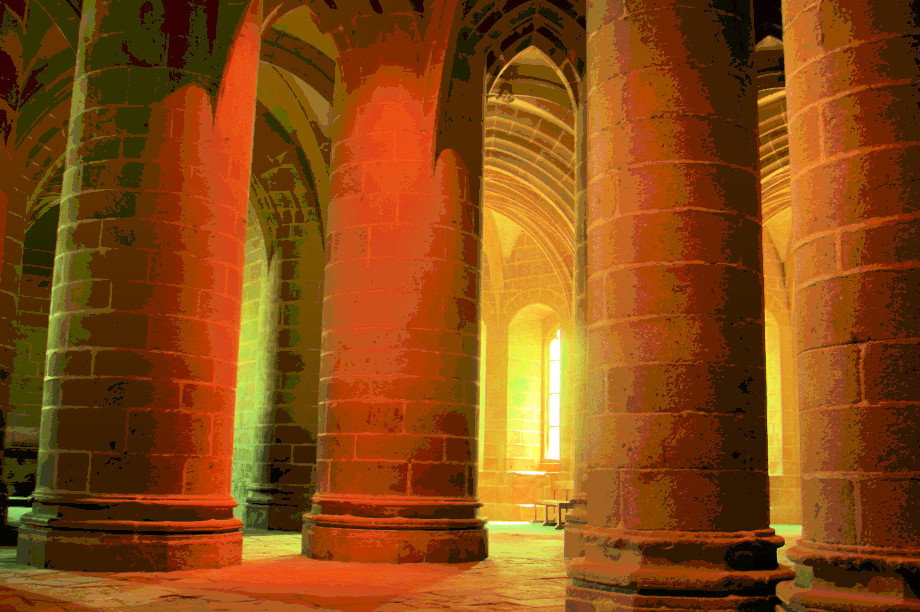}
    \\[2pt]
    \includegraphics[width=\textwidth, trim=0 150 0 0, clip]{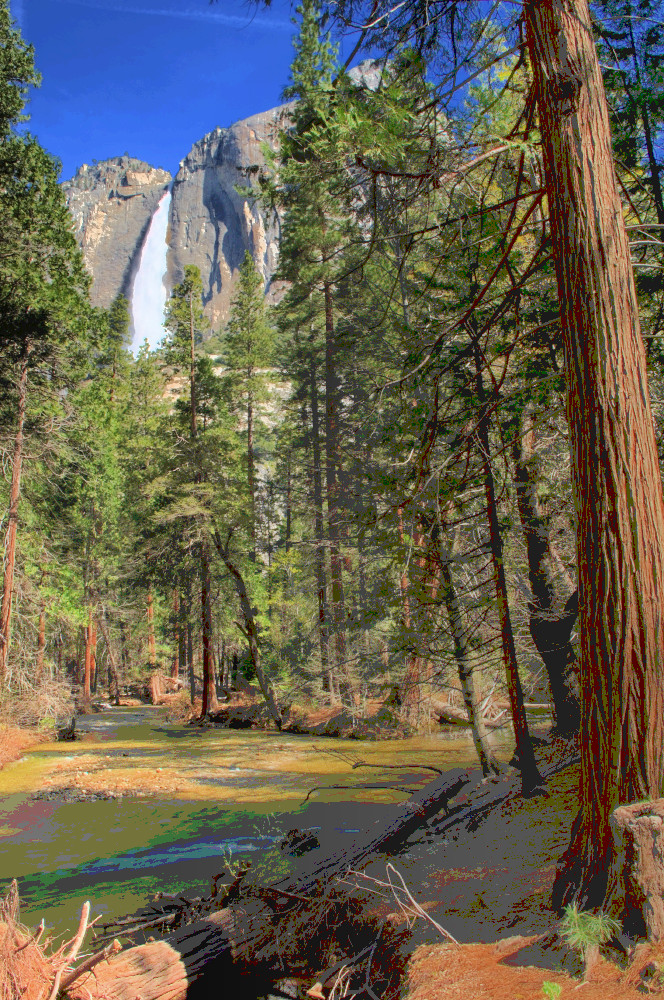}
    \caption{CLAHE}
  \end{subfigure}
  \begin{subfigure}[b]{\subfigwidth}
    \includegraphics[width=\textwidth, trim=0 0 0 0, clip]{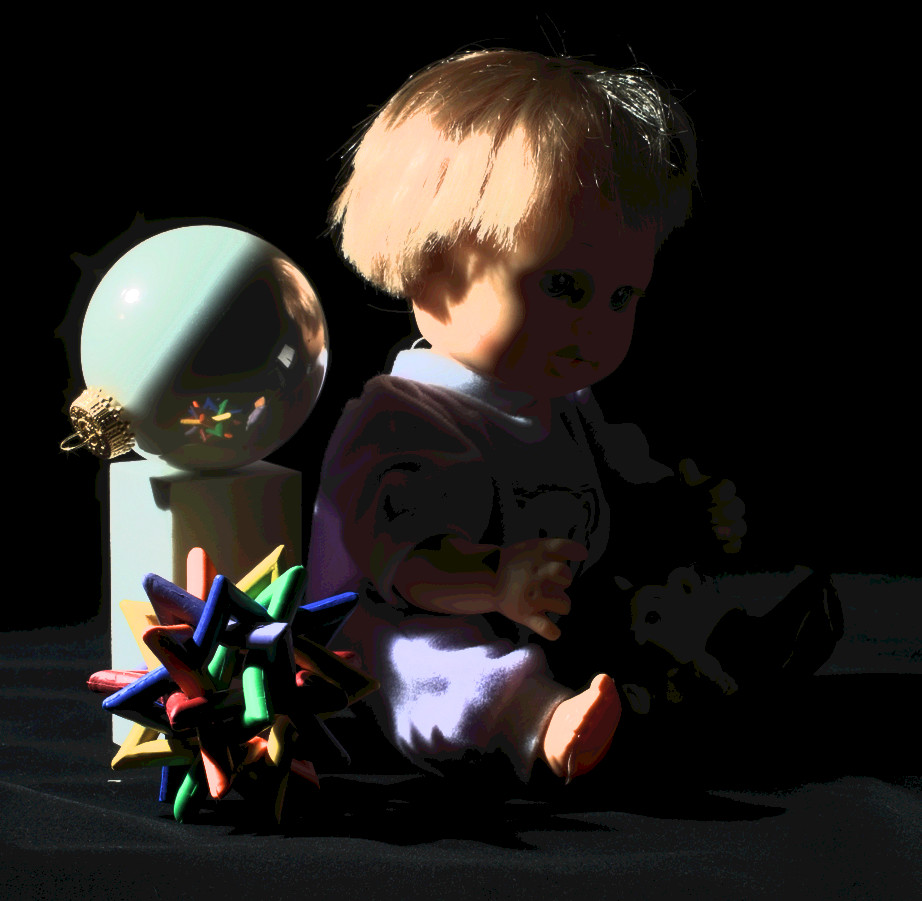}
    \\[2pt]
    \includegraphics[width=\textwidth, trim=50 0 150 0, clip]{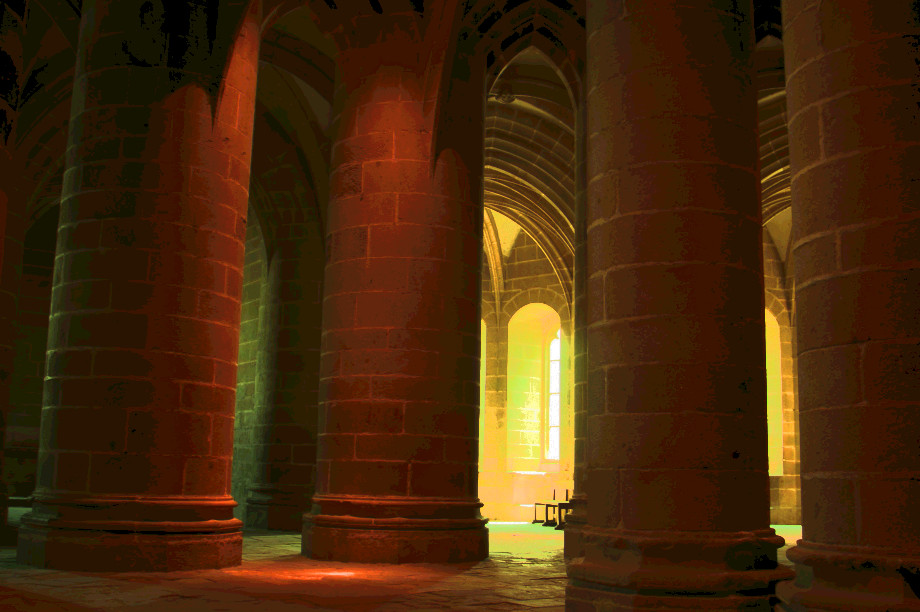}
    \\[2pt]
    \includegraphics[width=\textwidth, trim=0 150 0 0, clip]{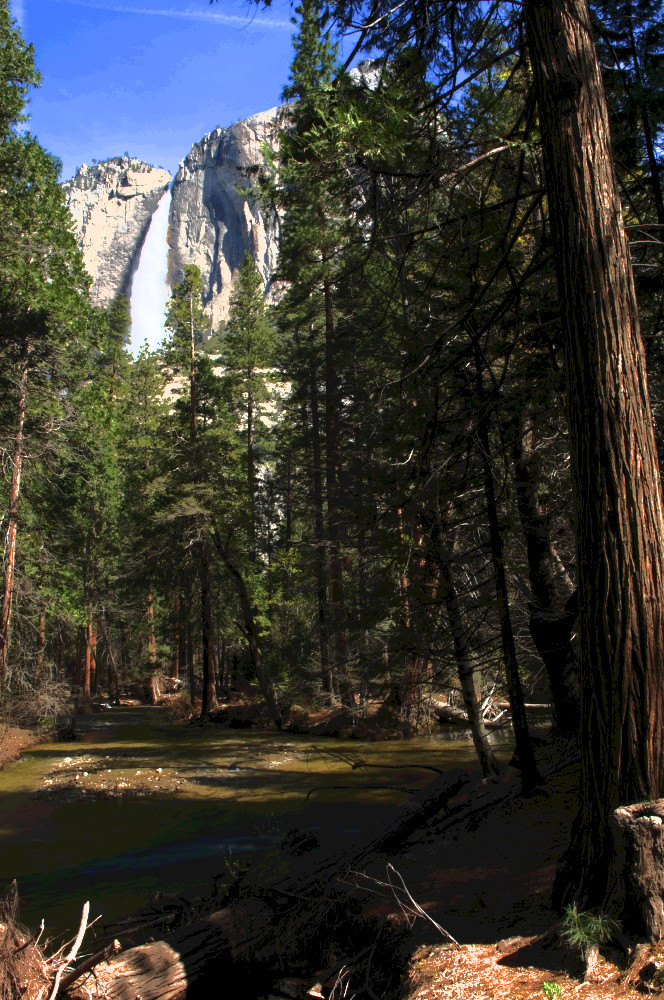}
    \caption{OCTM}
  \end{subfigure}
  \begin{subfigure}[b]{\subfigwidth}
    \includegraphics[width=\textwidth, trim=0 0 0 0, clip]{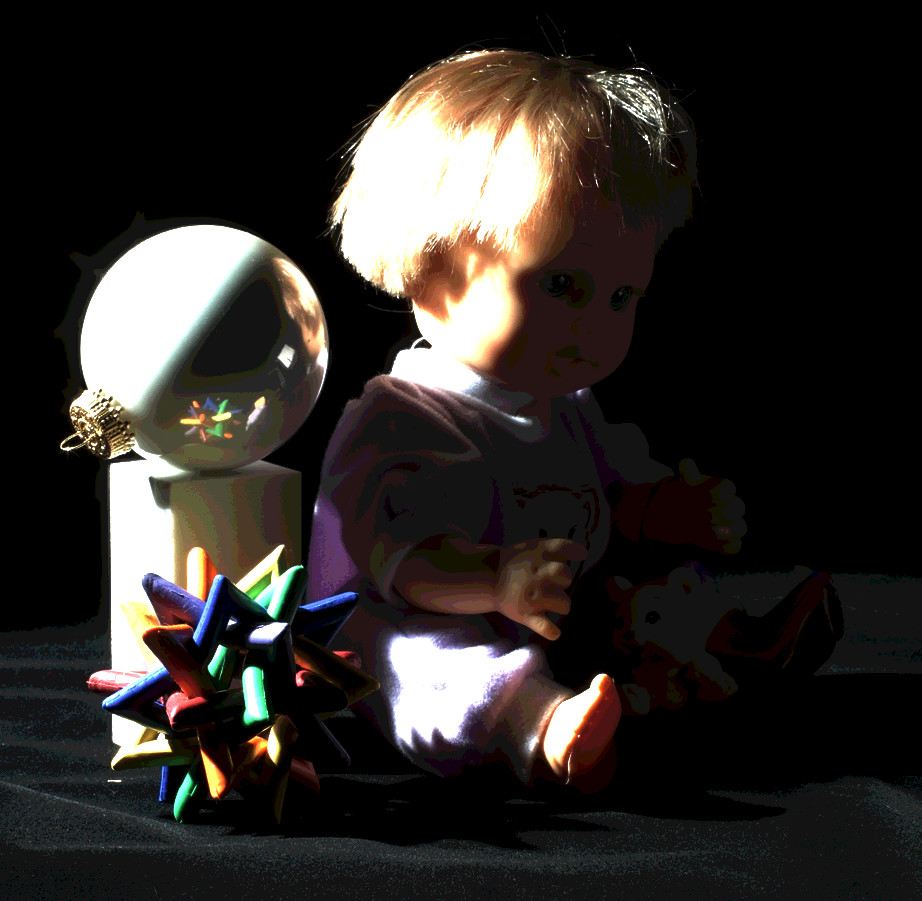}
    \\[2pt]
    \includegraphics[width=\textwidth, trim=50 0 150 0, clip]{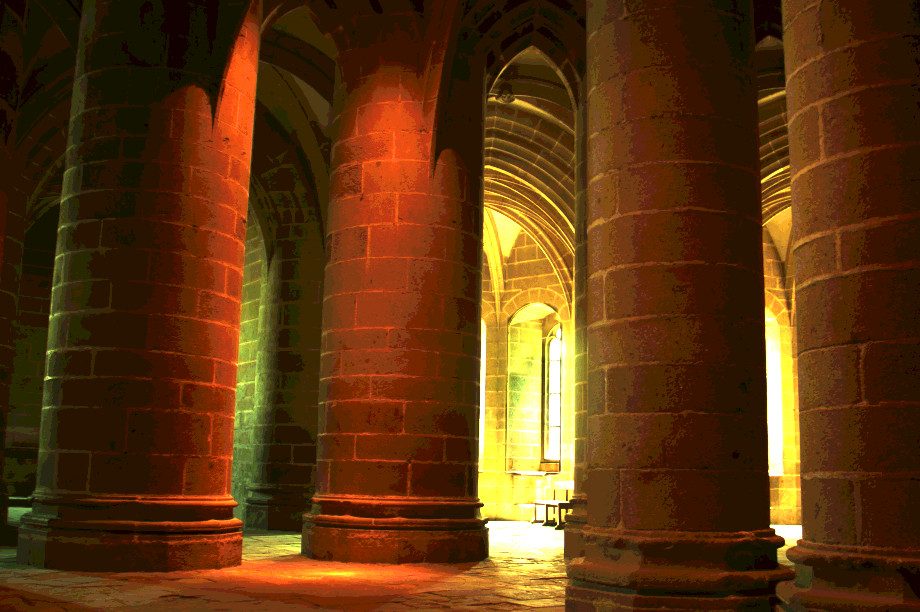}
    \\[2pt]
    \includegraphics[width=\textwidth, trim=0 150 0 0, clip]{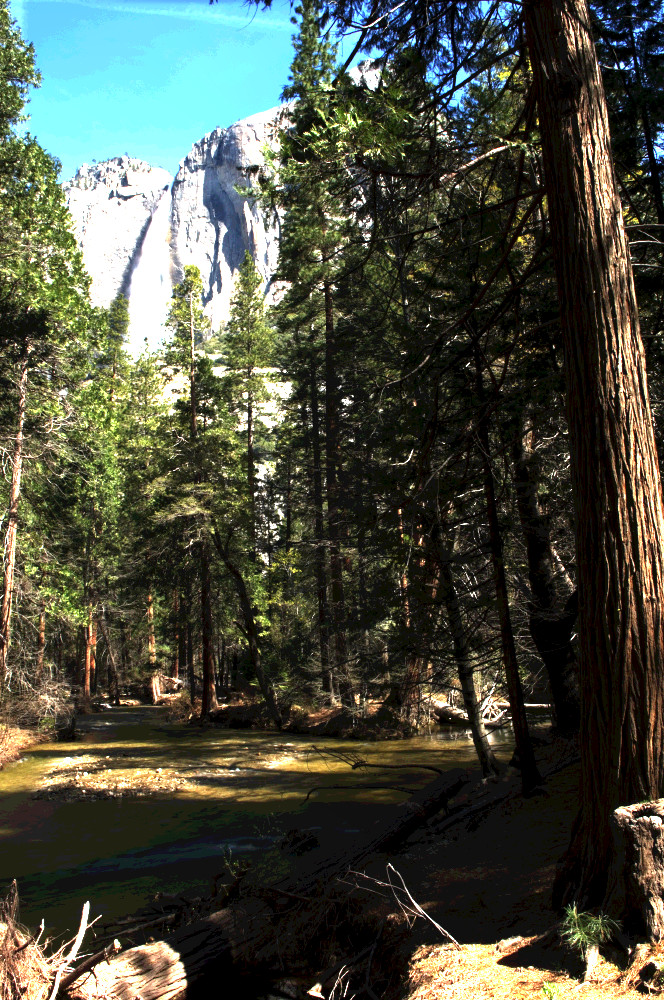}
    \caption{LIME}
  \end{subfigure}
  \begin{subfigure}[b]{\subfigwidth}
    \includegraphics[width=\textwidth, trim=0 0 0 0, clip]{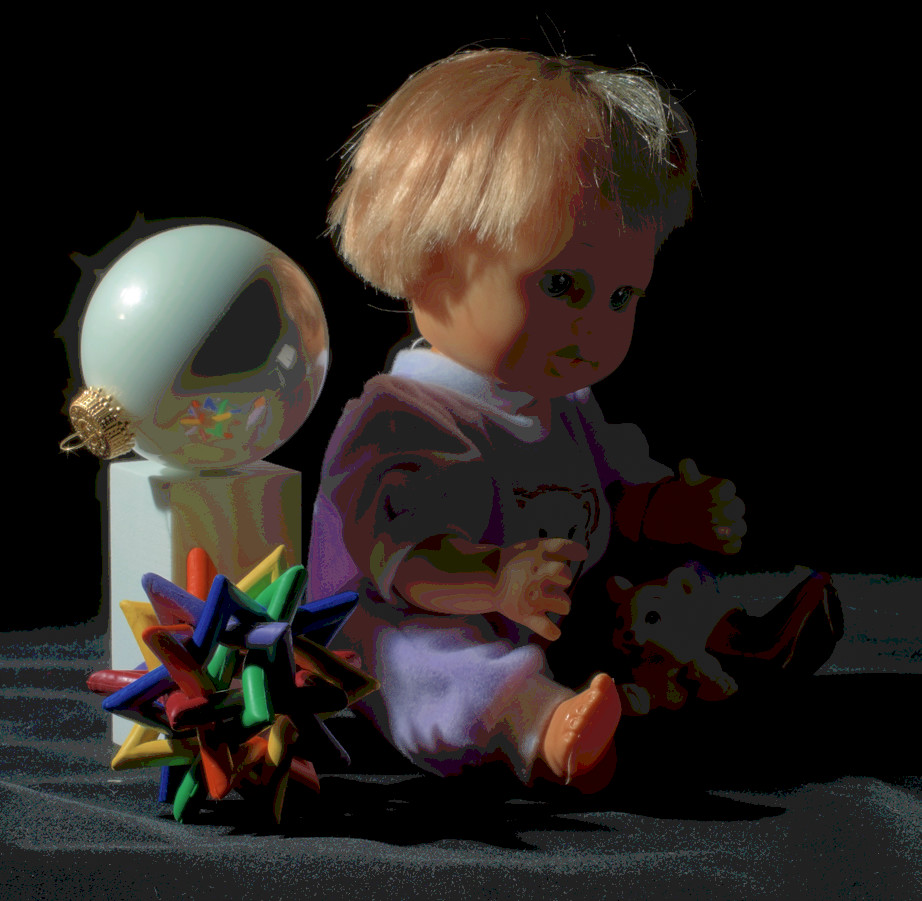}
    \\[2pt]
    \includegraphics[width=\textwidth, trim=50 0 150 0, clip]{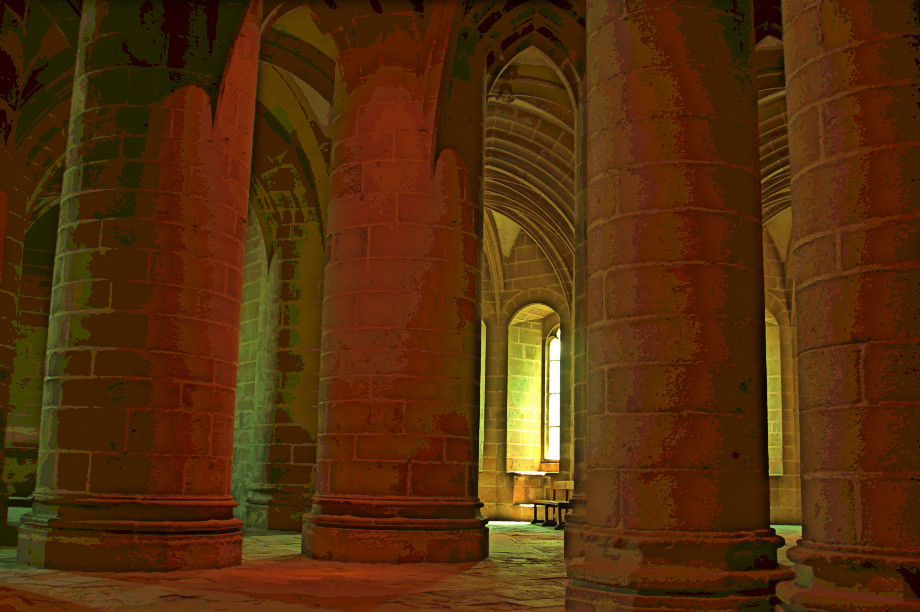}
    \\[2pt]
    \includegraphics[width=\textwidth, trim=0 150 0 0, clip]{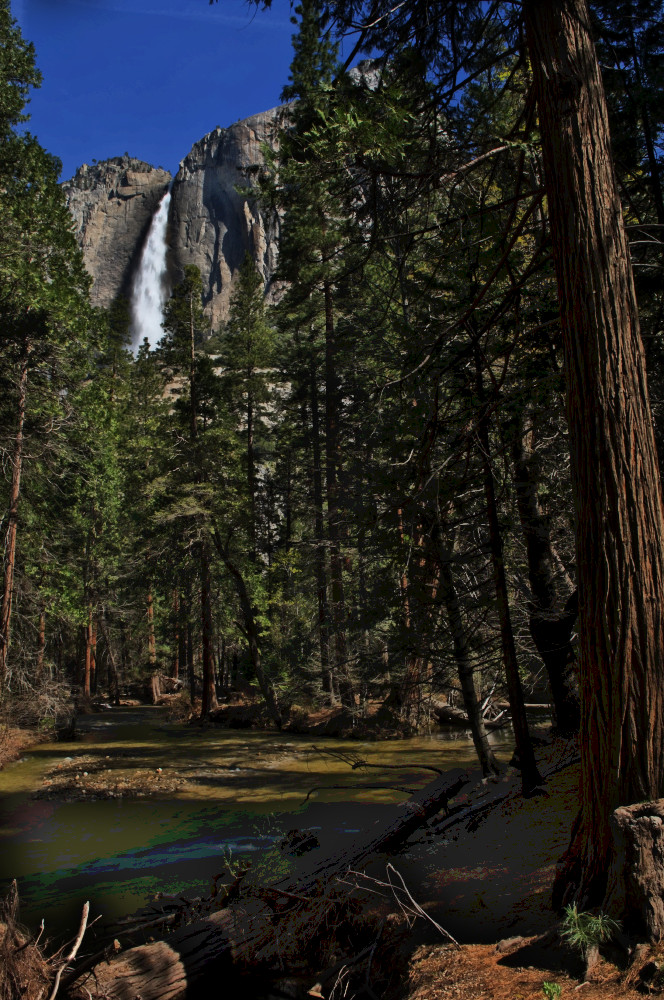}
    \caption{SRIE}
  \end{subfigure}
  \begin{subfigure}[b]{\subfigwidth}
    \includegraphics[width=\textwidth, trim=0 0 0 0, clip]{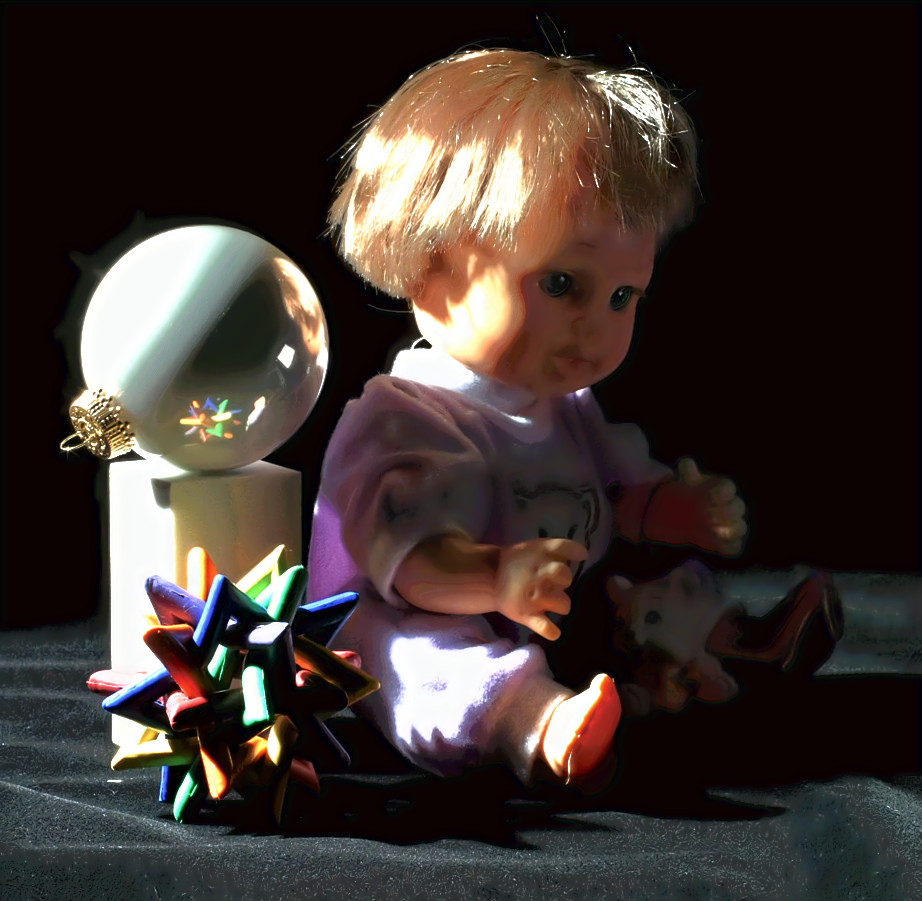}
    \\[2pt]
    \includegraphics[width=\textwidth, trim=50 0 150 0, clip]{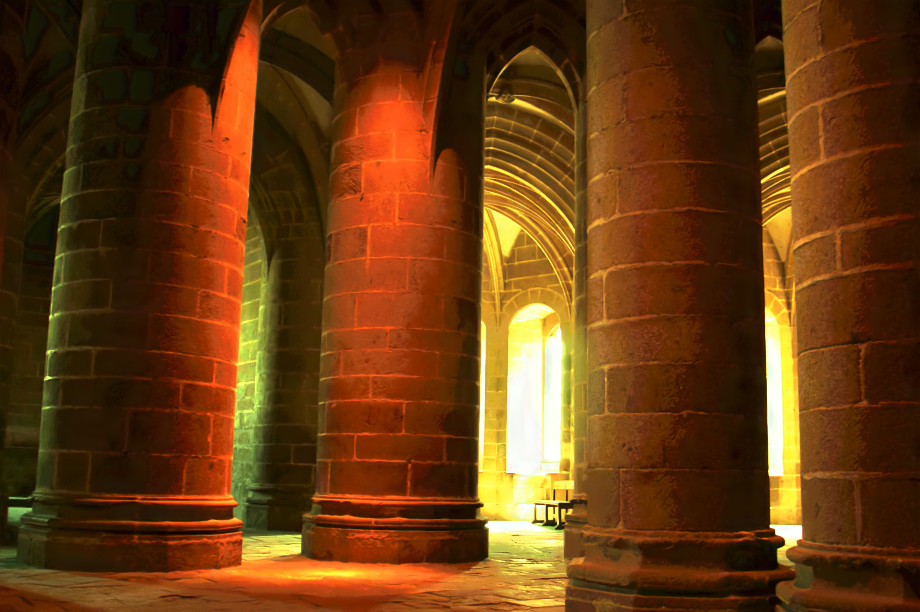}
    \\[2pt]
    \includegraphics[width=\textwidth, trim=0 150 0 0, clip]{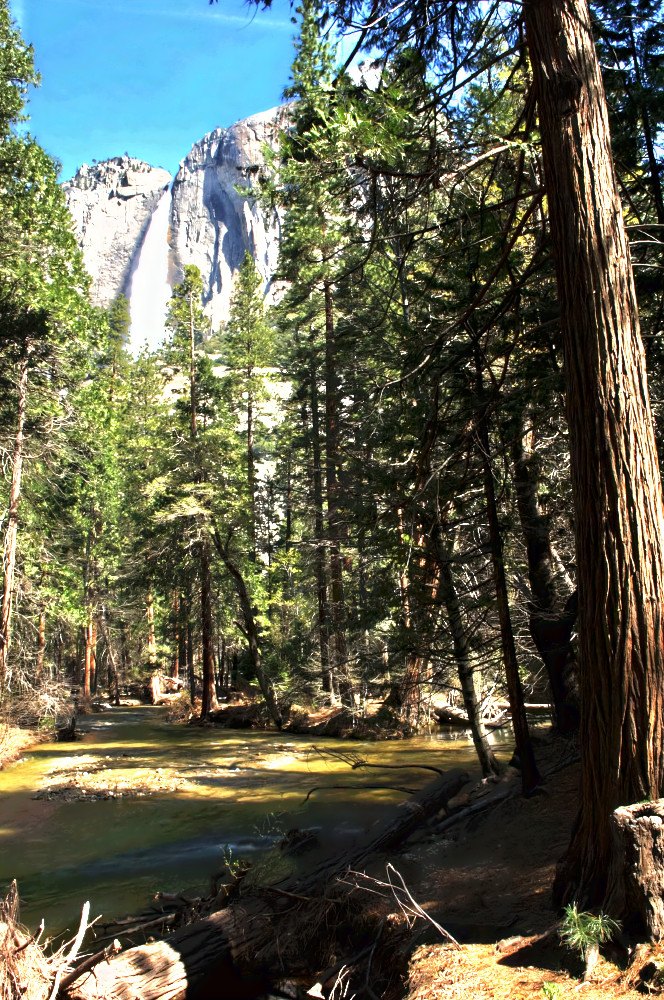}
    \caption{Proposed}
  \end{subfigure}
  \caption{Results of the tested techniques for low-light images
    synthesized from HDR images.}
  \label{fig:hdr-local}
\end{figure*}

Fig.~\ref{fig:syn-global} shows the visual comparison on two synthetic
global low light images.  The synthetic images are generated from some
normally exposed images from standard benchmark dataset
BSD500~\cite{martin2001database} by compressing their dynamic ranges
by the factor $a^{\gamma_2} = 1/30,\gamma_2/\gamma_1=1.3$.  Then random gaussian noise $n$($\mu=0,\sigma=0.25$) is
added to the images.  As shown by the figure, most tested techniques
adjust the lightness and contrast successfully, but only the proposed
technique can reduce the quantization artifact and restore some
missing details.

Fig.~\ref{fig:syn-local} shows another experiment on synthetic local
low-light images.  Dynamic range of the test images is scaled down by
the factor $a^{\gamma2} = 1/5,\gamma_2/\gamma_1=1.5$~\cite{guo2017lime}.  In these test results, there are obviously
contours in the results from the methods other than the proposed
technique, especially in the relative smooth area such as sky and
road.  Our method, in comparison, works well for the illumination
compensation and quantization residual restoration.  The test images
in Fig.~\ref{fig:hdr-local} are synthesized from high dynamic range
(HDR) images.  In this test case, the proposed algorithm still
outperforms the compared algorithms, although it has never been
trained using this type of synthetic low-light images.

\subsection{Experiments on Real Photographs}

\begin{figure*}[t]
  \centering

  \setlength{\subfigwidth}{0.49\linewidth}

  \begin{subfigure}[b]{\subfigwidth}
    \includegraphics[width=0.495\textwidth, trim=200 0 100 0, clip]{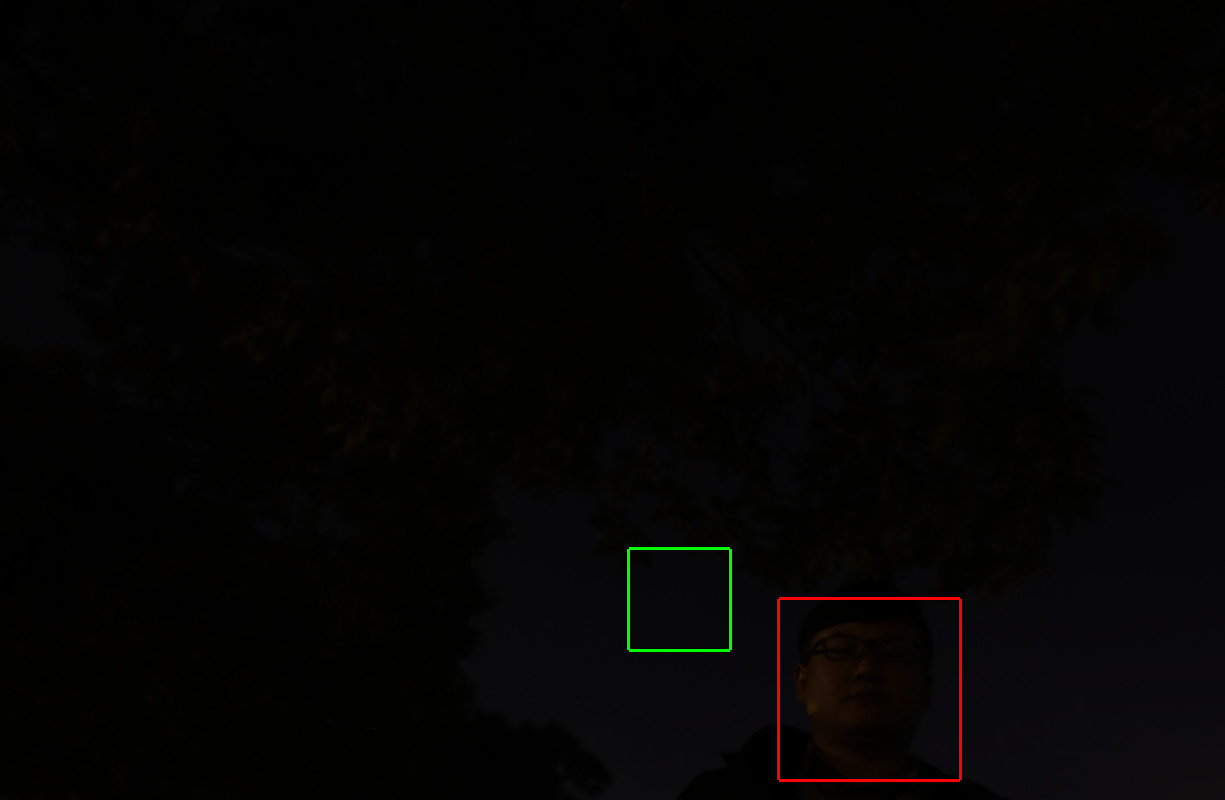}
    \includegraphics[width=0.495\textwidth, trim=172 0 50 0, clip]{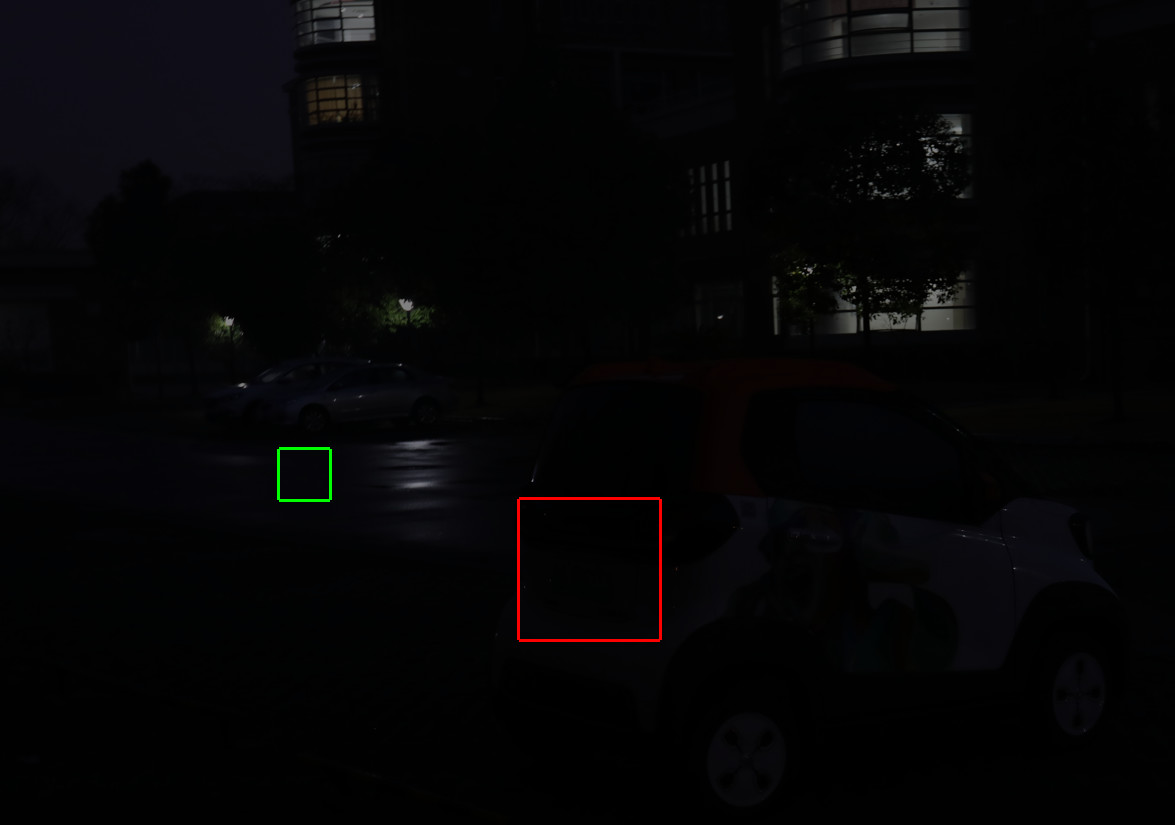}
    \includegraphics[width=0.243\textwidth]{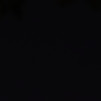}
    \hfill
    \includegraphics[width=0.243\textwidth]{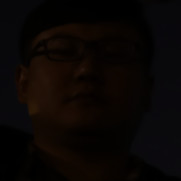}
    \includegraphics[width=0.243\textwidth]{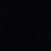}
    \hfill
    \includegraphics[width=0.243\textwidth]{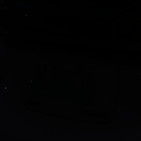}
    \caption{Input}
  \end{subfigure}
  \begin{subfigure}[b]{\subfigwidth}
    \includegraphics[width=0.495\textwidth, trim=200 0 100 0, clip]{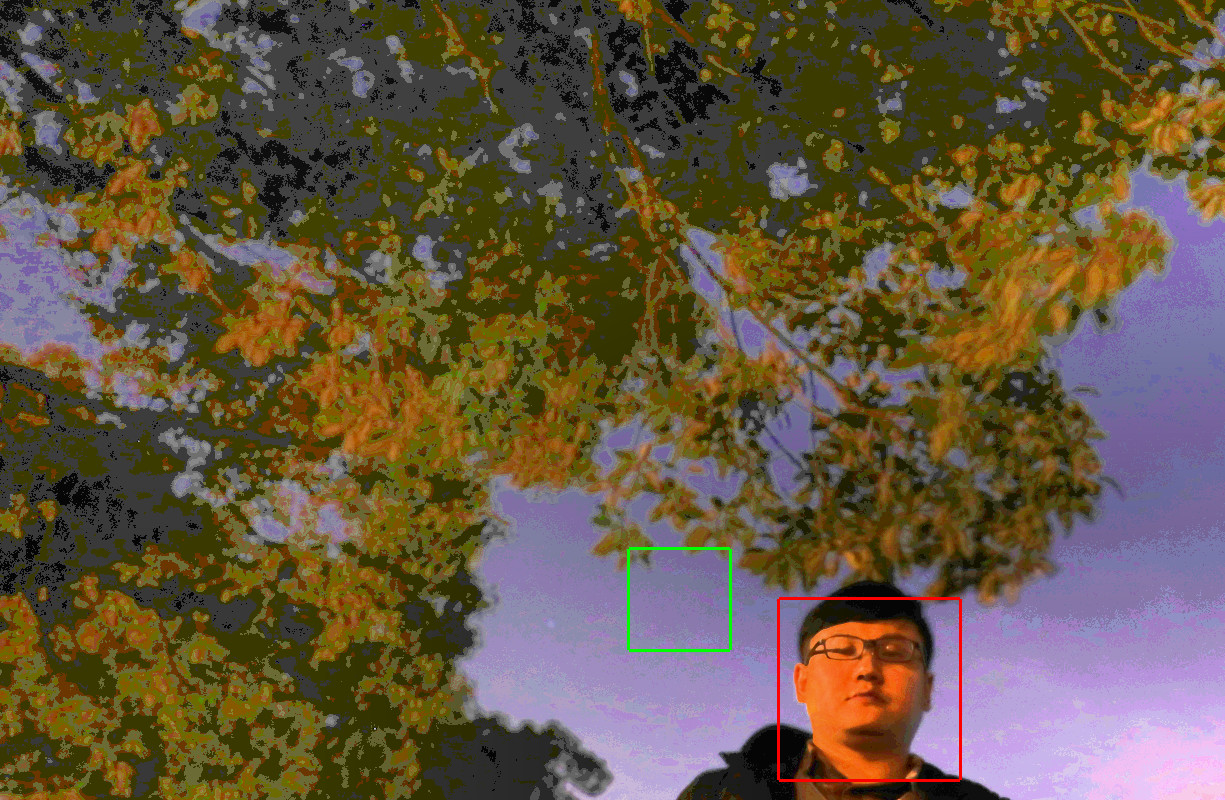}
    \includegraphics[width=0.495\textwidth, trim=172 0 50 0, clip]{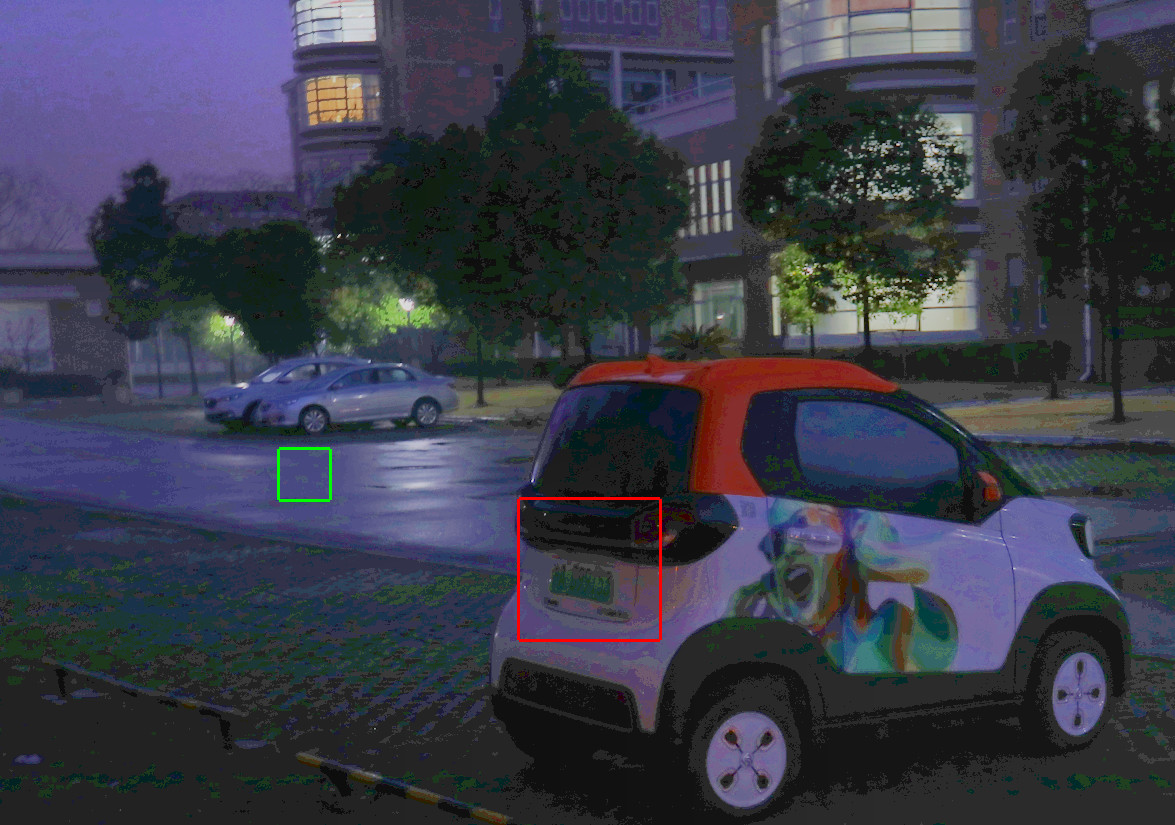}
    \includegraphics[width=0.243\textwidth]{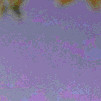}
    \hfill
    \includegraphics[width=0.243\textwidth]{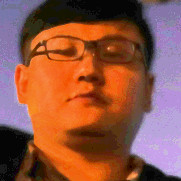}
    \includegraphics[width=0.243\textwidth]{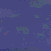}
    \hfill
    \includegraphics[width=0.243\textwidth]{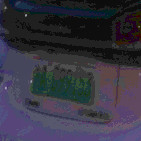}
    \caption{CLAHE}
  \end{subfigure}
  \begin{subfigure}[b]{\subfigwidth}
    \includegraphics[width=0.495\textwidth, trim=200 0 100 0, clip]{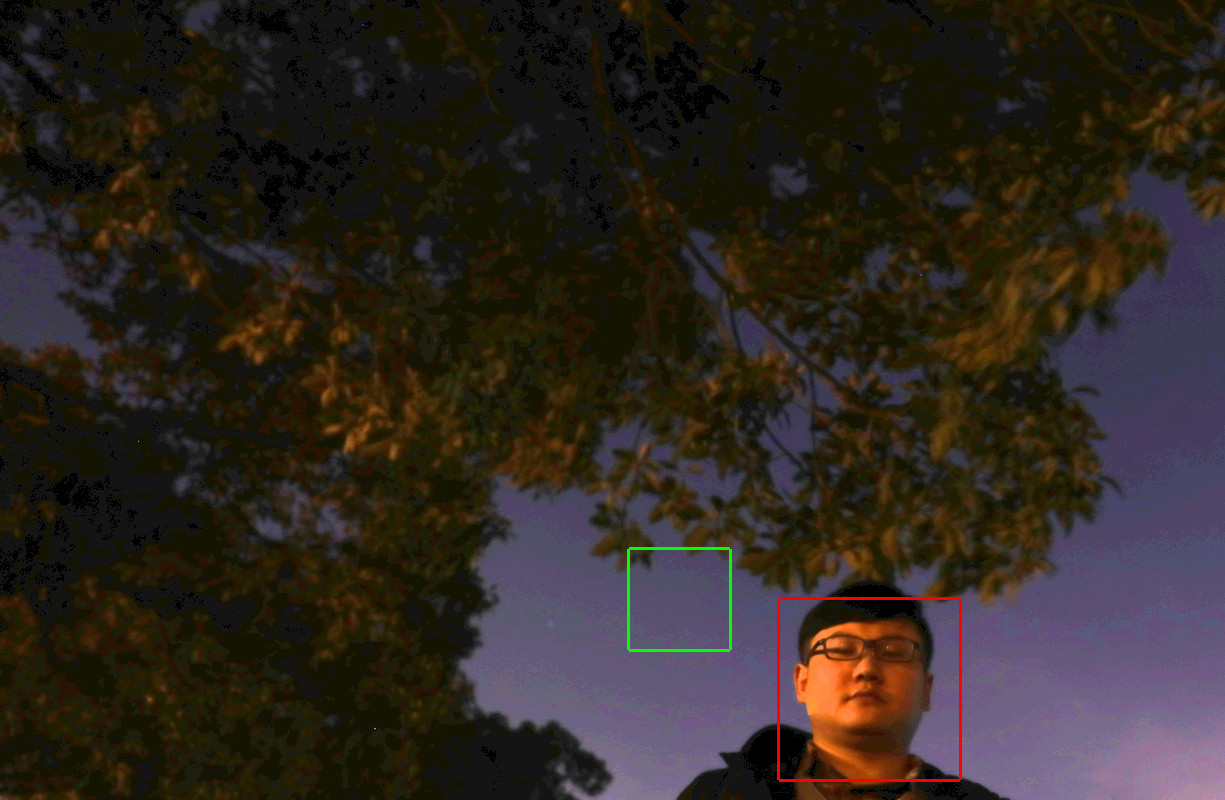}
    \includegraphics[width=0.495\textwidth, trim=172 0 50 0, clip]{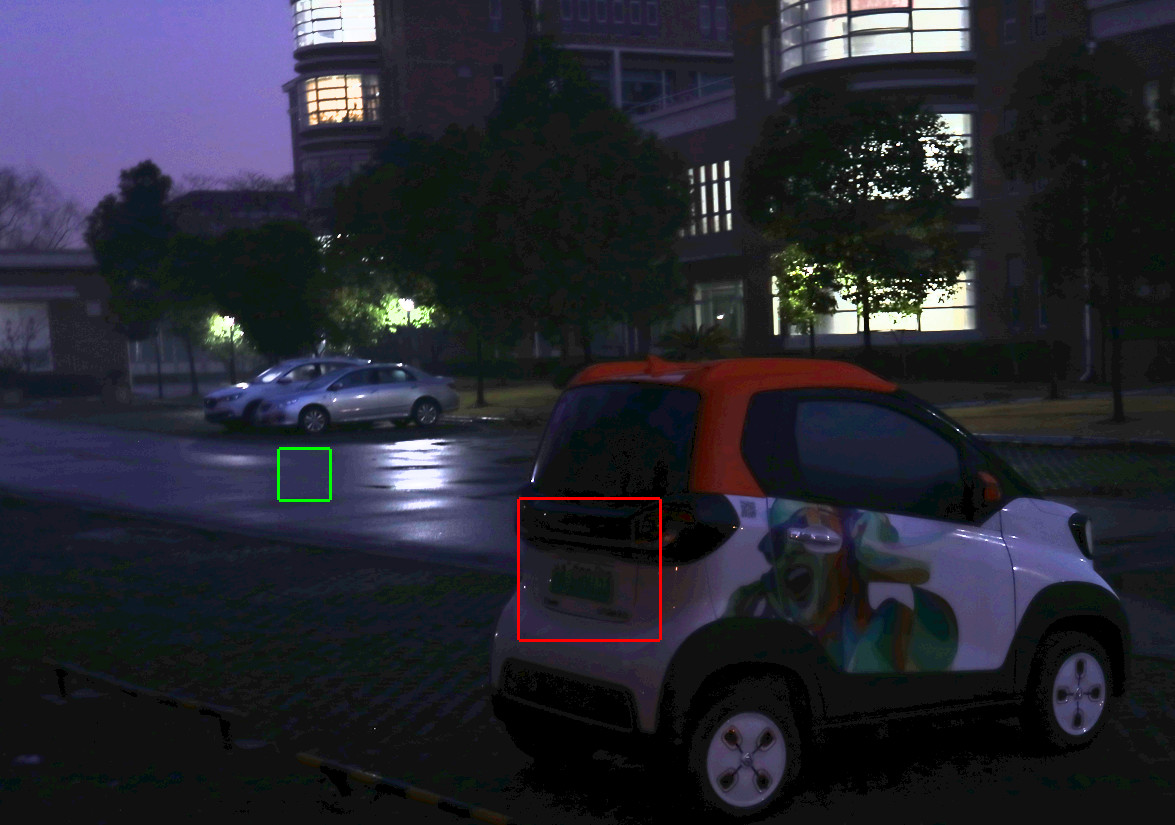}
    \includegraphics[width=0.243\textwidth]{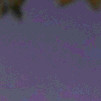}
    \hfill
    \includegraphics[width=0.243\textwidth]{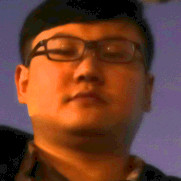}
    \includegraphics[width=0.243\textwidth]{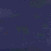}
    \hfill
    \includegraphics[width=0.243\textwidth]{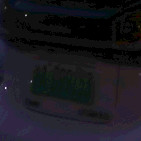}
    \caption{OCTM}
  \end{subfigure}
  \begin{subfigure}[b]{\subfigwidth}
    \includegraphics[width=0.495\textwidth, trim=200 0 100 0, clip]{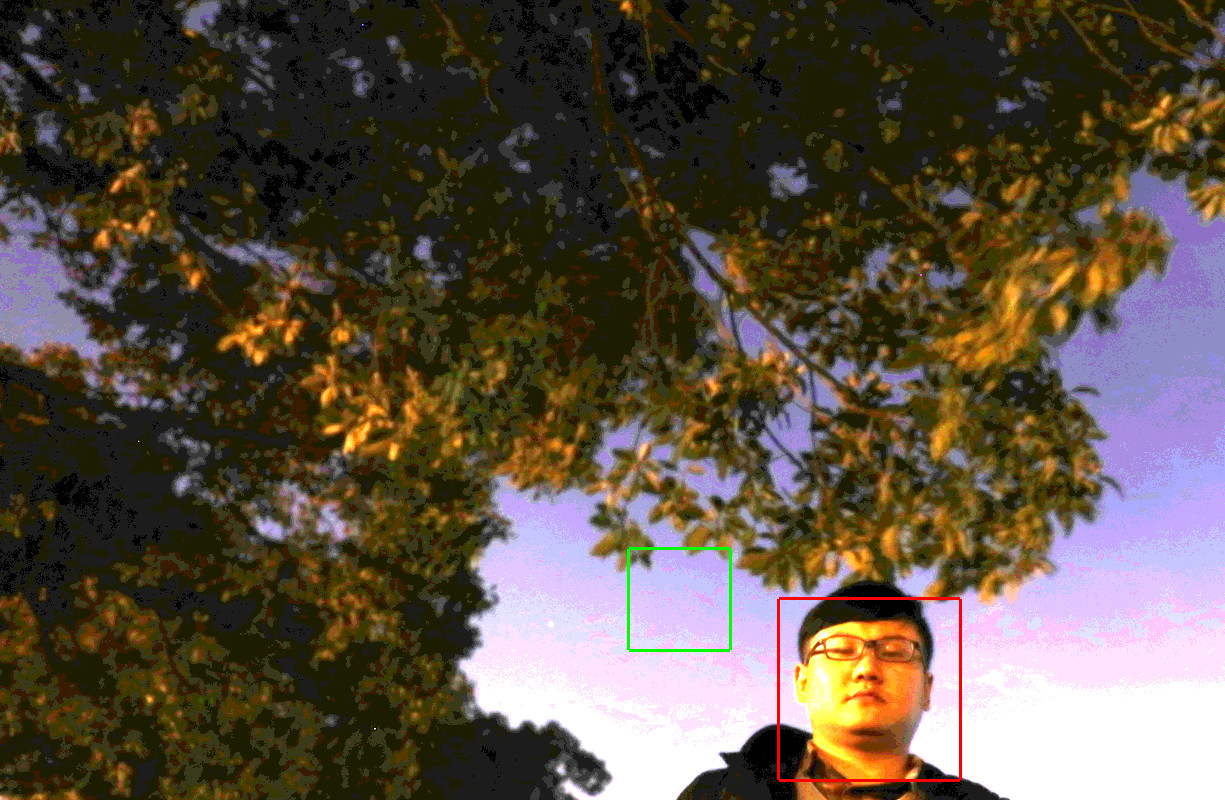}
    \includegraphics[width=0.495\textwidth, trim=172 0 50 0, clip]{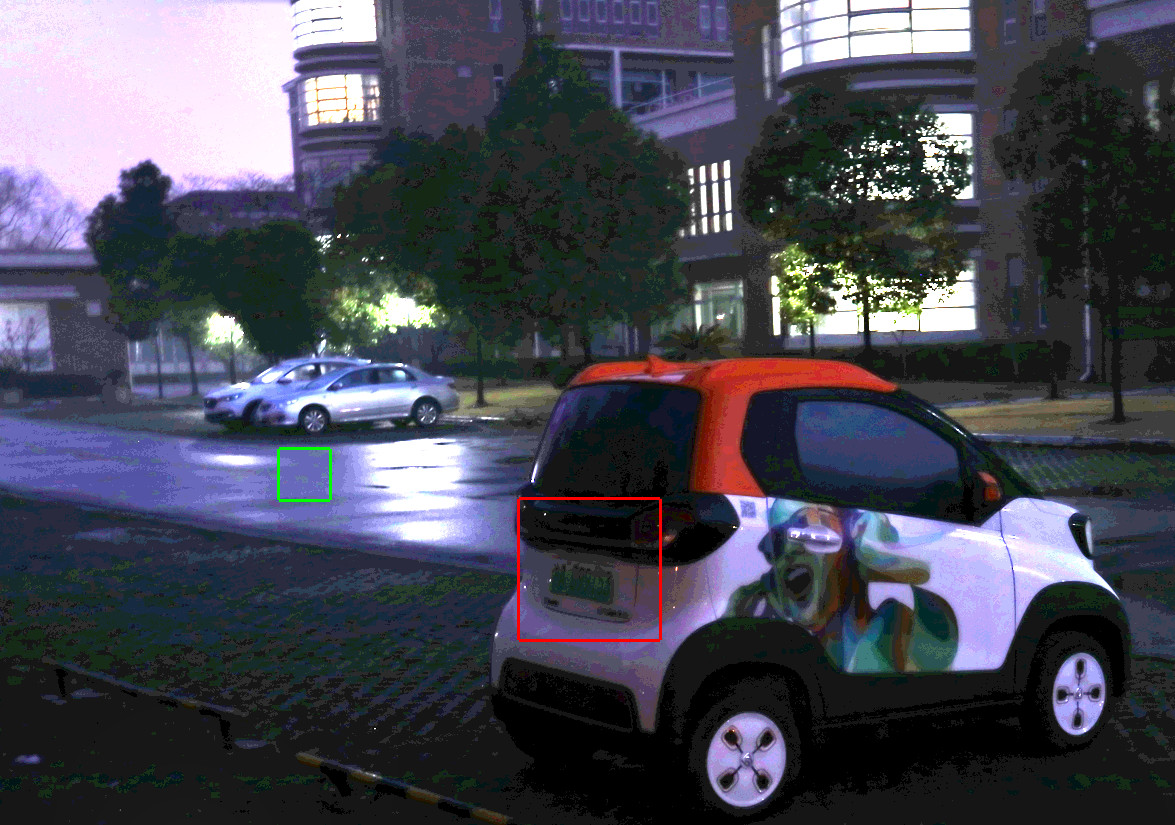}
    \includegraphics[width=0.243\textwidth]{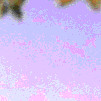}
    \hfill
    \includegraphics[width=0.243\textwidth]{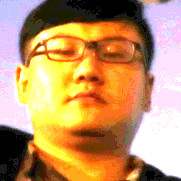}
    \includegraphics[width=0.243\textwidth]{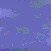}
    \hfill
    \includegraphics[width=0.243\textwidth]{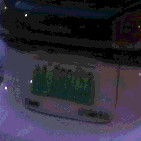}
    \caption{LIME}
  \end{subfigure}
  \begin{subfigure}[b]{\subfigwidth}
    \includegraphics[width=0.495\textwidth, trim=200 0 100 0, clip]{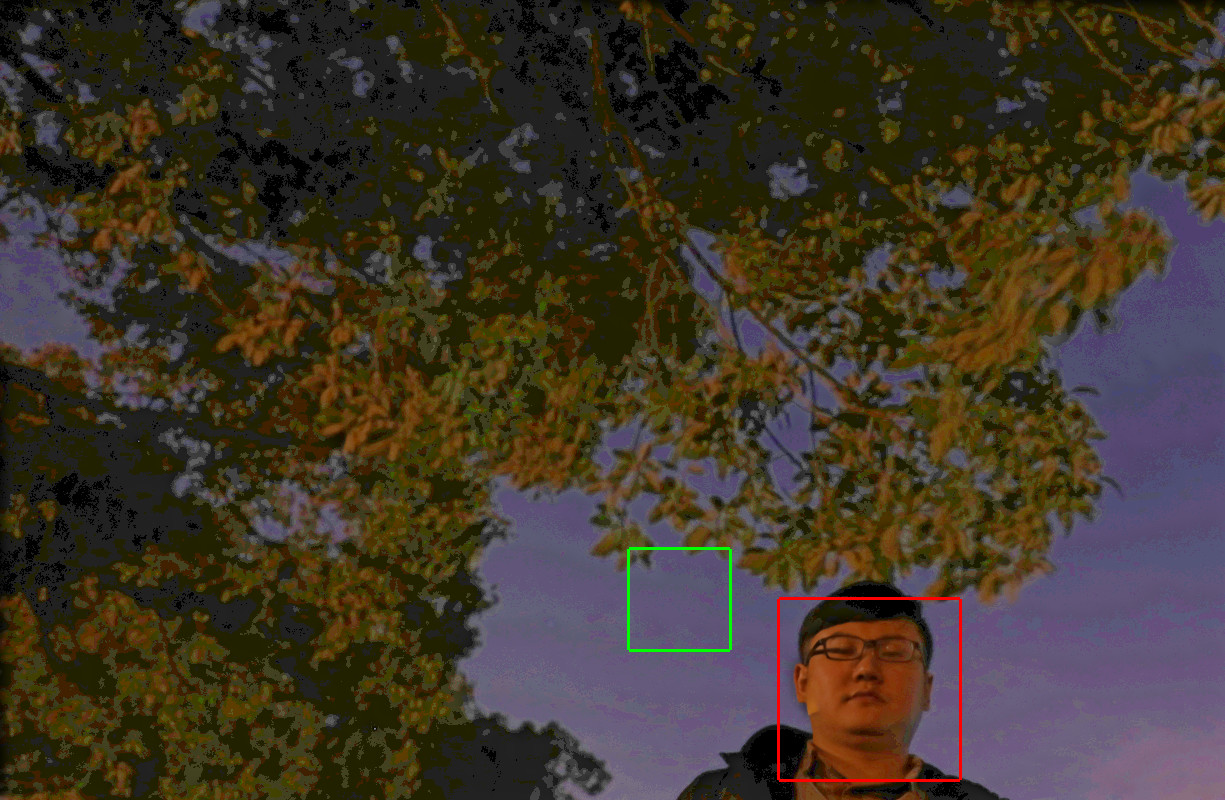}
    \includegraphics[width=0.495\textwidth, trim=172 0 50 0, clip]{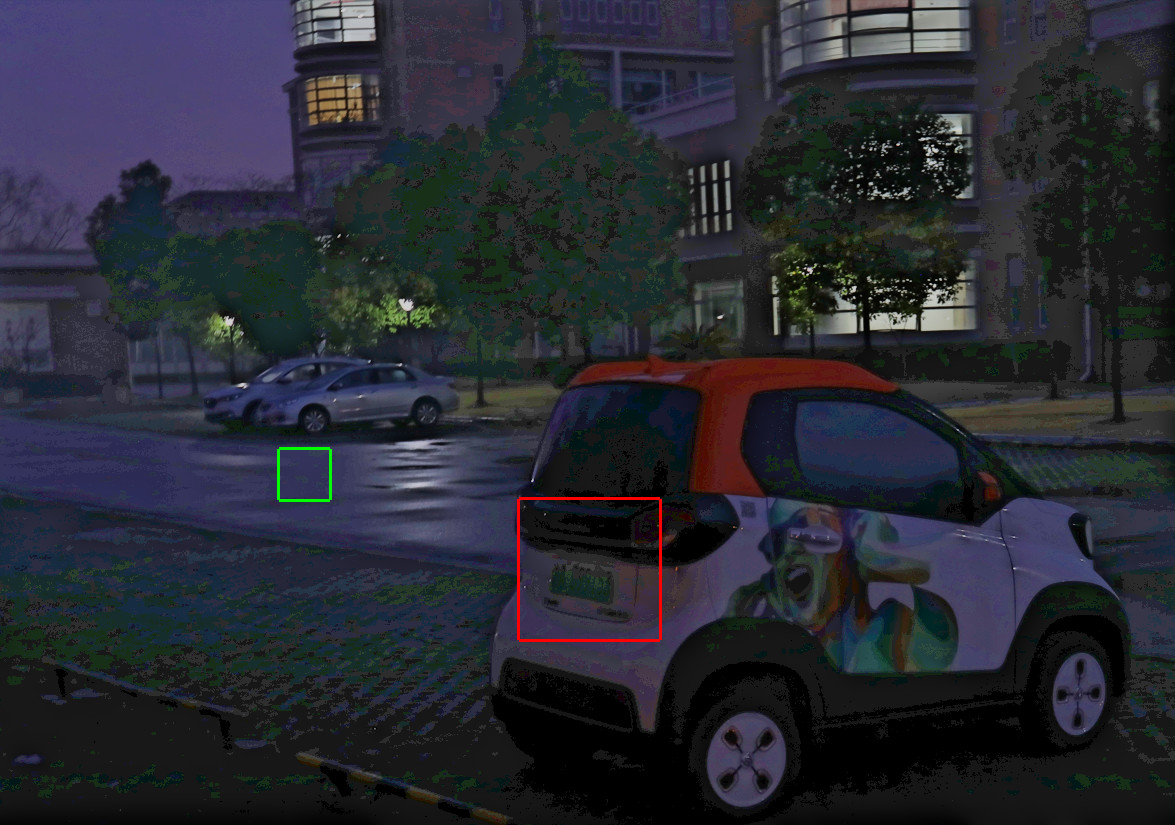}
    \includegraphics[width=0.243\textwidth]{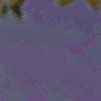}
    \hfill
    \includegraphics[width=0.243\textwidth]{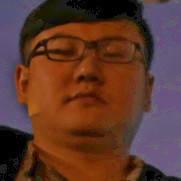}
    \includegraphics[width=0.243\textwidth]{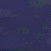}
    \hfill
    \includegraphics[width=0.243\textwidth]{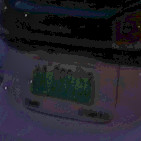}
    \caption{SRIE}
  \end{subfigure}
  \begin{subfigure}[b]{\subfigwidth}
    \includegraphics[width=0.495\textwidth, trim=200 0 100 0, clip]{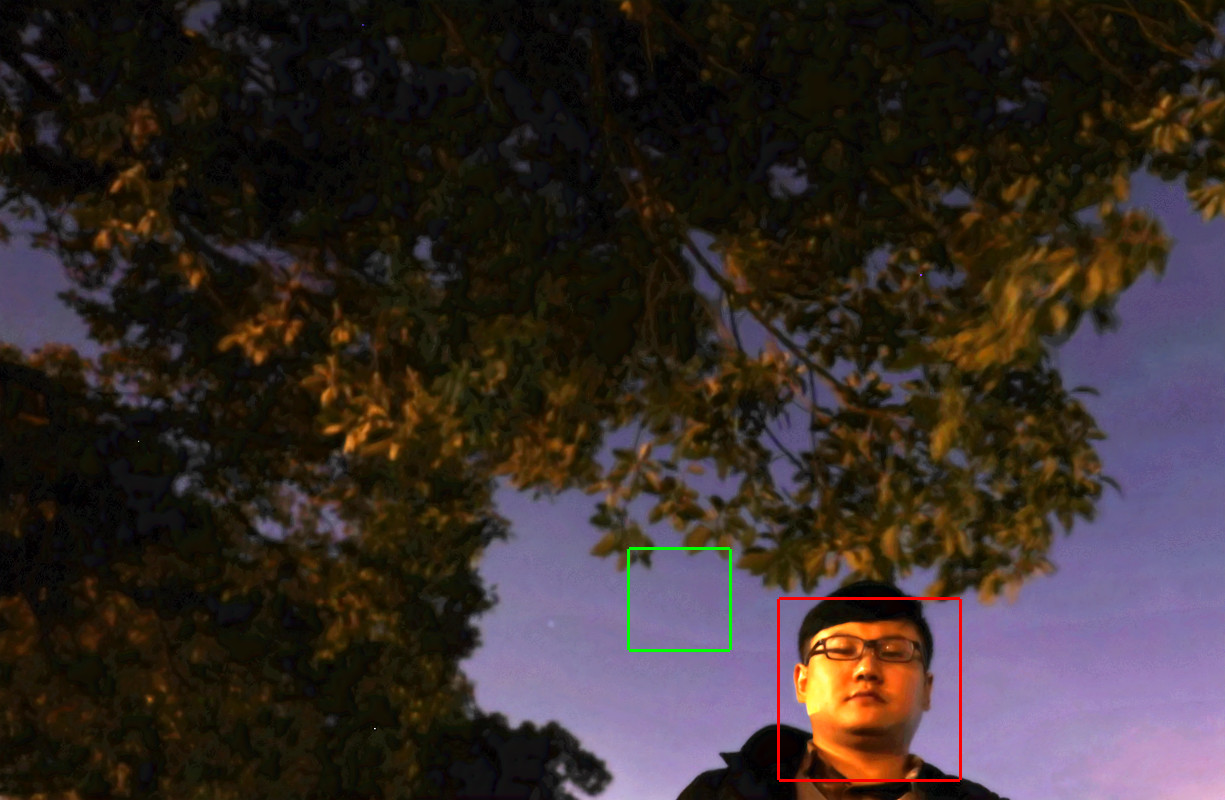}
    \includegraphics[width=0.495\textwidth, trim=172 0 50 0, clip]{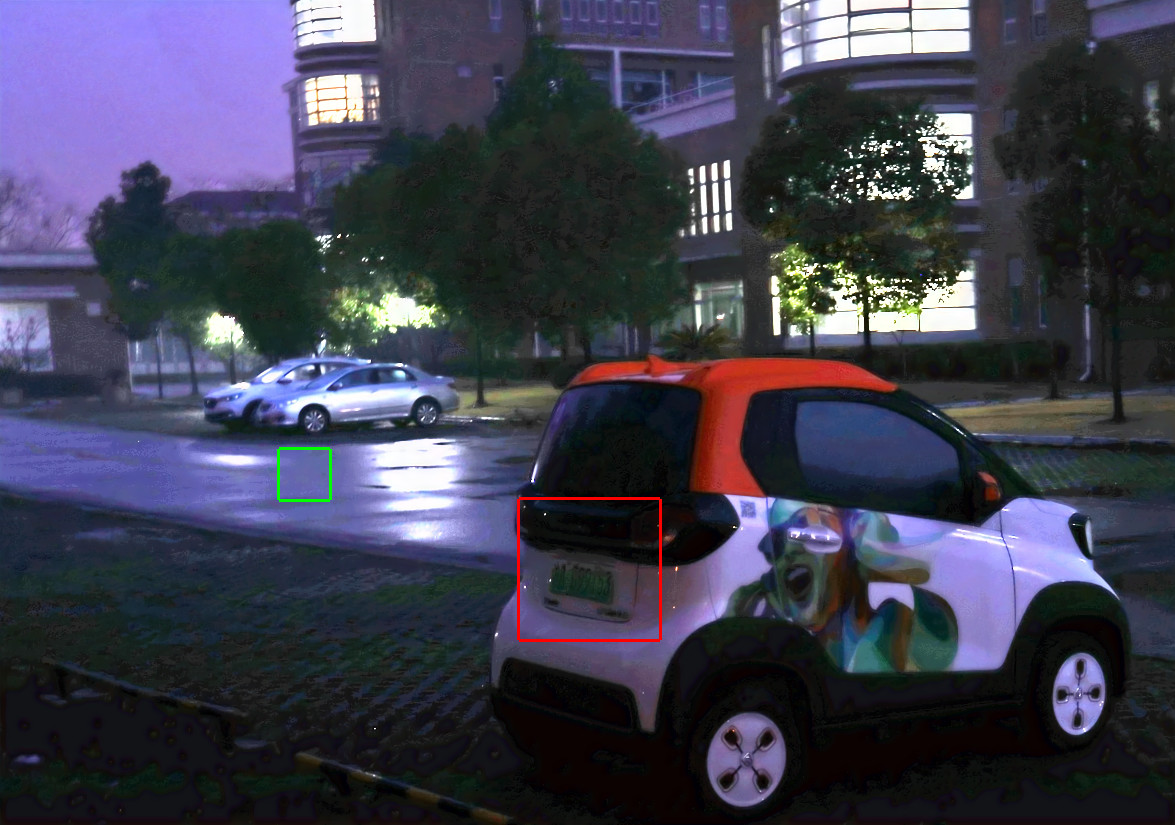}
    \includegraphics[width=0.243\textwidth]{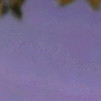}
    \hfill
    \includegraphics[width=0.243\textwidth]{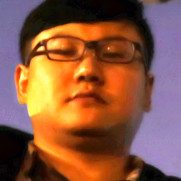}
    \includegraphics[width=0.243\textwidth]{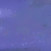}
    \hfill
    \includegraphics[width=0.243\textwidth]{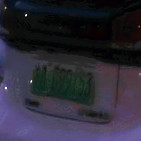}
    \caption{Proposed}
  \end{subfigure}

  \caption{Results of the tested techniques for real photographs.}
  \label{fig:real-detail}
\end{figure*}

Experiments on synthetic images demonstrates the superiority of the
propose method.  For real photographs, the proposed algorithm also
works well.  Fig.~\ref{fig:real-detail} shows some examples of real
photographs.  Comparing the performance of the five methods, CLAHE can
effectively increase the illuminance, but often resulting in low
contrast.  LIME over-enhances the image.  OCTM and SRIE performs
better, but their results are not so impressive.  For better
evaluation, we can see the details in the green and red square.
Blocking artifacts appears in all the images enhanced by the four
other methods, except for our method.  Our method also enhance the
human face better than the other techniques and restore the license
plate to some extent.


\bibliographystyle{IEEEbib}
\bibliography{dequant}

\begin{thebibliography}{10}

\bibitem{ciregan2012multi}
Dan Ciregan, Ueli Meier, and J{\"u}rgen Schmidhuber,
\newblock ``Multi-column deep neural networks for image classification,''
\newblock in {\em Computer Vision and Pattern Recognition (CVPR), 2012 IEEE
  Conference on}. IEEE, 2012, pp. 3642--3649.

\bibitem{yang2009linear}
Jianchao Yang, Kai Yu, Yihong Gong, and Thomas Huang,
\newblock ``Linear spatial pyramid matching using sparse coding for image
  classification,''
\newblock in {\em Computer Vision and Pattern Recognition, 2009. CVPR 2009.
  IEEE Conference on}. IEEE, 2009, pp. 1794--1801.

\bibitem{he2016deep}
Kaiming He, Xiangyu Zhang, Shaoqing Ren, and Jian Sun,
\newblock ``Deep residual learning for image recognition,''
\newblock in {\em Proceedings of the IEEE conference on computer vision and
  pattern recognition}, 2016, pp. 770--778.

\bibitem{simonyan2014very}
Karen Simonyan and Andrew Zisserman,
\newblock ``Very deep convolutional networks for large-scale image
  recognition,''
\newblock {\em arXiv preprint arXiv:1409.1556}, 2014.

\bibitem{yilmaz2006object}
Alper Yilmaz, Omar Javed, and Mubarak Shah,
\newblock ``Object tracking: A survey,''
\newblock {\em Acm computing surveys (CSUR)}, vol. 38, no. 4, pp. 13, 2006.

\bibitem{pisano1998contrast}
Etta~D Pisano, Shuquan Zong, Bradley~M Hemminger, Marla DeLuca, R~Eugene
  Johnston, Keith Muller, M~Patricia Braeuning, and Stephen~M Pizer,
\newblock ``Contrast limited adaptive histogram equalization image processing
  to improve the detection of simulated spiculations in dense mammograms,''
\newblock {\em Journal of Digital imaging}, vol. 11, no. 4, pp. 193--200, 1998.

\bibitem{abdullah2007dynamic}
Mohammad Abdullah-Al-Wadud, Md~Hasanul Kabir, M~Ali~Akber Dewan, and Oksam
  Chae,
\newblock ``A dynamic histogram equalization for image contrast enhancement,''
\newblock {\em IEEE Transactions on Consumer Electronics}, vol. 53, no. 2,
  2007.

\bibitem{celik2011contextual}
Turgay Celik and Tardi Tjahjadi,
\newblock ``Contextual and variational contrast enhancement,''
\newblock {\em IEEE Transactions on Image Processing}, vol. 20, no. 12, pp.
  3431--3441, 2011.

\bibitem{lee2013contrast}
Chulwoo Lee, Chul Lee, and Chang-Su Kim,
\newblock ``Contrast enhancement based on layered difference representation of
  2d histograms,''
\newblock {\em IEEE Transactions on Image Processing}, vol. 22, no. 12, pp.
  5372--5384, 2013.

\bibitem{wu2011linear}
Xiaolin Wu,
\newblock ``A linear programming approach for optimal contrast-tone mapping,''
\newblock {\em IEEE transactions on image processing}, vol. 20, no. 5, pp.
  1262--1272, 2011.

\bibitem{li2014contrast}
Zhenhao Li and Xiaolin Wu,
\newblock ``Contrast enhancement with chromaticity error bound,''
\newblock in {\em Image Processing (ICIP), 2014 IEEE International Conference
  on}. IEEE, 2014, pp. 4507--4511.

\bibitem{land1977retinex}
Edwin~H Land,
\newblock ``The retinex theory of color vision,''
\newblock {\em Scientific American}, vol. 237, no. 6, pp. 108--129, 1977.

\bibitem{jobson1997properties}
Daniel~J Jobson, Zia-ur Rahman, and Glenn~A Woodell,
\newblock ``Properties and performance of a center/surround retinex,''
\newblock {\em IEEE transactions on image processing}, vol. 6, no. 3, pp.
  451--462, 1997.

\bibitem{jobson1997multiscale}
Daniel~J Jobson, Zia-ur Rahman, and Glenn~A Woodell,
\newblock ``A multiscale retinex for bridging the gap between color images and
  the human observation of scenes,''
\newblock {\em IEEE Transactions on Image processing}, vol. 6, no. 7, pp.
  965--976, 1997.

\bibitem{guo2017lime}
Xiaojie Guo, Yu~Li, and Haibin Ling,
\newblock ``Lime: Low-light image enhancement via illumination map
  estimation,''
\newblock {\em IEEE Transactions on Image Processing}, vol. 26, no. 2, pp.
  982--993, 2017.

\bibitem{fu2016weighted}
Xueyang Fu, Delu Zeng, Yue Huang, Xiao-Ping Zhang, and Xinghao Ding,
\newblock ``A weighted variational model for simultaneous reflectance and
  illumination estimation,''
\newblock in {\em Proceedings of the IEEE Conference on Computer Vision and
  Pattern Recognition}, 2016, pp. 2782--2790.

\bibitem{dong2011fast}
Xuan Dong, Guan Wang, Yi~Pang, Weixin Li, Jiangtao Wen, Wei Meng, and Yao Lu,
\newblock ``Fast efficient algorithm for enhancement of low lighting video,''
\newblock in {\em Multimedia and Expo (ICME), 2011 IEEE International
  Conference on}. IEEE, 2011, pp. 1--6.

\bibitem{li2015low}
Lin Li, Ronggang Wang, Wenmin Wang, and Wen Gao,
\newblock ``A low-light image enhancement method for both denoising and
  contrast enlarging,''
\newblock in {\em Image Processing (ICIP), 2015 IEEE International Conference
  on}. IEEE, 2015, pp. 3730--3734.

\bibitem{loza2013automatic}
Artur {\L}oza, David~R Bull, Paul~R Hill, and Alin~M Achim,
\newblock ``Automatic contrast enhancement of low-light images based on local
  statistics of wavelet coefficients,''
\newblock {\em Digital Signal Processing}, vol. 23, no. 6, pp. 1856--1866,
  2013.

\bibitem{lore2017llnet}
Kin~Gwn Lore, Adedotun Akintayo, and Soumik Sarkar,
\newblock ``Llnet: A deep autoencoder approach to natural low-light image
  enhancement,''
\newblock {\em Pattern Recognition}, vol. 61, pp. 650--662, 2017.

\bibitem{ioffe2015batch}
Sergey Ioffe and Christian Szegedy,
\newblock ``Batch normalization: Accelerating deep network training by reducing
  internal covariate shift,''
\newblock in {\em International Conference on Machine Learning}, 2015, pp.
  448--456.

\bibitem{radford2015unsupervised}
Alec Radford, Luke Metz, and Soumith Chintala,
\newblock ``Unsupervised representation learning with deep convolutional
  generative adversarial networks,''
\newblock {\em arXiv preprint arXiv:1511.06434}, 2015.

\bibitem{nair2010rectified}
Vinod Nair and Geoffrey~E Hinton,
\newblock ``Rectified linear units improve restricted boltzmann machines,''
\newblock in {\em Proceedings of the 27th international conference on machine
  learning (ICML-10)}, 2010, pp. 807--814.

\bibitem{goodfellow2014generative}
Ian Goodfellow, Jean Pouget-Abadie, Mehdi Mirza, Bing Xu, David Warde-Farley,
  Sherjil Ozair, Aaron Courville, and Yoshua Bengio,
\newblock ``Generative adversarial nets,''
\newblock in {\em Advances in neural information processing systems}, 2014, pp.
  2672--2680.

\bibitem{ledig2016photo}
Christian Ledig, Lucas Theis, Ferenc Husz{\'a}r, Jose Caballero, Andrew
  Cunningham, Alejandro Acosta, Andrew Aitken, Alykhan Tejani, Johannes Totz,
  Zehan Wang, et~al.,
\newblock ``Photo-realistic single image super-resolution using a generative
  adversarial network,''
\newblock {\em arXiv preprint arXiv:1609.04802}, 2016.

\bibitem{guo2016one}
Jun Guo and Hongyang Chao,
\newblock ``One-to-many network for visually pleasing compression artifacts
  reduction,''
\newblock {\em arXiv preprint arXiv:1611.04994}, 2016.

\bibitem{zfarbman2008}
Zeev Farbman, Raanan Fattal, Dani Lischinski, and Richard Szeliski,
\newblock ``Edge-preserving decompositions for multi-scale tone and detail
  manipulation,''
\newblock {\em ACM Transactions on Graphics}, vol. 27, no. 3, pp. 67:1--67:10,
  2008.

\bibitem{rfattal2002}
Raanan Fattal, Dani Lischinski, and Michael Werman,
\newblock ``Gradient domain high dynamic range compression,''
\newblock {\em ACM Transactions on Graphics}, vol. 21, no. 3, pp. 249--256,
  2002.

\bibitem{sparis2011}
Sylvain Paris, Samuel~W. Hasinoff, and Jan Kautz,
\newblock ``Local laplacian filters: Edge-aware image processing with a
  laplacian pyramid,''
\newblock {\em ACM Transactions on Graphics}, vol. 30, no. 4, pp. 68:1--68:12,
  2011.

\bibitem{zrahman2004}
Ziaur Rahman and Glenn~A. Woodell,
\newblock ``Retinex processing for automatic image enhancement,''
\newblock {\em Journal of Electronic Imaging}, vol. 13, pp. 100--110, 2004.

\bibitem{ereinhard2002}
Erik Reinhard, Michael Stark, Peter Shirley, and James Ferwerda,
\newblock ``Photographic tone reproduction for digital images,''
\newblock {\em ACM Transactions on Graphics}, vol. 21, no. 3, pp. 267--276,
  2002.

\bibitem{martin2001database}
David Martin, Charless Fowlkes, Doron Tal, and Jitendra Malik,
\newblock ``A database of human segmented natural images and its application to
  evaluating segmentation algorithms and measuring ecological statistics,''
\newblock in {\em Computer Vision, 2001. ICCV 2001. Proceedings. Eighth IEEE
  International Conference on}. IEEE, 2001, vol.~2, pp. 416--423.

\end{thebibliography}

\end{document}